\documentclass[11pt, logo, copyright, colorlinks=true, allcolors=blue]{nvidiatechreport}

\PassOptionsToPackage{numbers, compress}{natbib}
\usepackage[utf8]{inputenc} 
\usepackage[T1]{fontenc}    
\usepackage{url}            
\usepackage{booktabs}       
\usepackage{amsfonts}       
\usepackage{nicefrac}       
\usepackage{microtype}      
\usepackage{xcolor}         
\usepackage{colortbl}       
\usepackage{multirow}       
\usepackage[numbers]{natbib}  

\usepackage{bbm}
\usepackage{pifont}
\usepackage{subcaption}
\usepackage{graphicx}

\usepackage{amsmath,amssymb,amsthm}
\usepackage{booktabs}
\usepackage{enumitem}
\usepackage[most]{tcolorbox}
\usepackage{amsmath, amssymb, amsthm}
\usepackage{geometry}
\usepackage{mathtools}
\usepackage{xspace}
\usepackage{dsfont}
\usepackage{fvextra}
\usepackage{hyperref}

\newtheorem{proposition}{Proposition}
\theoremstyle{definition}

\theoremstyle{remark}

\definecolor{remarkgreen}{HTML}{2d6a4f}
\definecolor{claimred}{HTML}{c25e3f}
\definecolor{takeawayblue}{HTML}{1f4e8a}
\definecolor{Red}{rgb}{0.768, 0.054, 0.054}
\definecolor{Blue}{rgb}{0.152, 0.294, 0.925}
\definecolor{Green}{rgb}{0,0.4,0.7}
\definecolor{darkgray}{gray}{0.3}
\definecolor{chocolate}{HTML}{D2691E}
\definecolor{maroon}{HTML}{A00000}
\definecolor{indigo}{HTML}{4B0082}
\definecolor{green}{HTML}{008000}
\definecolor{newred}{HTML}{a91e1e}
\definecolor{cadmiumgreen}{rgb}{0.0, 0.42, 0.24}
\definecolor{forestgreen}{rgb}{0.13, 0.55, 0.13}
\definecolor{deltateal}{HTML}{0F766E}
\definecolor{mygray}{gray}{0.92}
\definecolor{myblue}{HTML}{E8F0FB}
\definecolor{linkcolor}{RGB}{190,105,30}

\newtcolorbox{remarkbox}{
  colback=remarkgreen!5, colframe=remarkgreen!40,
  fonttitle=\bfseries, left=4pt, right=4pt, top=4pt, bottom=4pt,
  boxrule=0pt, leftrule=2pt, sharp corners
}
\newtcolorbox{claimbox}{
  colback=claimred!5, colframe=claimred!30,
  fonttitle=\bfseries, left=4pt, right=4pt, top=4pt, bottom=4pt,
  boxrule=0.5pt, sharp corners
}
\tcbset{
  takeawaybox/.style={
    width=\linewidth,
    boxrule=0.8pt,
    top=6pt, bottom=2pt, left=3pt, right=3pt,
    before skip=12pt, after skip=8pt,
    colback=takeawayblue!3!white,
    colframe=takeawayblue,
    colbacktitle=takeawayblue,
    enhanced, breakable,
    attach boxed title to top left={yshift=-0.1in, xshift=0.15in},
    boxed title style={boxrule=0pt, colframe=white},
  }
}
\newtcolorbox{takeawaybox}[2][]{takeawaybox, title=#2, #1}

\definecolor{axpoboxframe}{HTML}{2c5fa8}
\definecolor{grpoboxframe}{HTML}{a91e1e}
\definecolor{qboxframe}{HTML}{4a4a4a}
\tcbset{
  questionbox/.style={
    enhanced, breakable, sharp corners,
    width=\linewidth,
    boxrule=0.5pt, leftrule=2.5pt,
    top=4pt, bottom=4pt, left=6pt, right=6pt,
    before skip=4pt, after skip=4pt,
    colback=qboxframe!4!white,
    colframe=qboxframe!60,
    colbacktitle=qboxframe,
    coltitle=white,
    fonttitle=\bfseries\small,
    attach boxed title to top left={yshift=-2pt, xshift=4pt},
    boxed title style={boxrule=0pt, sharp corners, top=2pt, bottom=2pt, left=4pt, right=4pt},
  },
  axpobox/.style={
    enhanced, breakable, sharp corners,
    width=\linewidth,
    boxrule=0.5pt, leftrule=2.5pt,
    top=4pt, bottom=4pt, left=6pt, right=6pt,
    before skip=4pt, after skip=4pt,
    colback=axpoboxframe!4!white,
    colframe=axpoboxframe!60,
    colbacktitle=axpoboxframe,
    coltitle=white,
    fonttitle=\bfseries\small,
    attach boxed title to top left={yshift=-2pt, xshift=4pt},
    boxed title style={boxrule=0pt, sharp corners, top=2pt, bottom=2pt, left=4pt, right=4pt},
  },
  grpobox/.style={
    enhanced, breakable, sharp corners,
    width=\linewidth,
    boxrule=0.5pt, leftrule=2.5pt,
    top=4pt, bottom=4pt, left=6pt, right=6pt,
    before skip=4pt, after skip=4pt,
    colback=grpoboxframe!4!white,
    colframe=grpoboxframe!60,
    colbacktitle=grpoboxframe,
    coltitle=white,
    fonttitle=\bfseries\small,
    attach boxed title to top left={yshift=-2pt, xshift=4pt},
    boxed title style={boxrule=0pt, sharp corners, top=2pt, bottom=2pt, left=4pt, right=4pt},
  },
}
\newtcolorbox{questionbox}[1][]{questionbox, title=Question, #1}
\newtcolorbox{axpobox}[1][]{axpobox, title=AXPO\ \(\checkmark\), #1}
\newtcolorbox{grpobox}[1][]{grpobox, title=GRPO\ \(\times\), #1}

\tcbset{
  promptbox/.style={
    enhanced, breakable, sharp corners,
    width=\linewidth,
    boxrule=0.5pt, leftrule=2.5pt,
    top=4pt, bottom=4pt, left=6pt, right=6pt,
    before skip=4pt, after skip=4pt,
    colback=qboxframe!4!white,
    colframe=qboxframe!60,
    colbacktitle=qboxframe,
    coltitle=white,
    fonttitle=\bfseries\small,
    attach boxed title to top left={yshift=-2pt, xshift=4pt},
    boxed title style={boxrule=0pt, sharp corners, top=2pt, bottom=2pt, left=4pt, right=4pt},
  },
}
\newtcolorbox{promptbox}[1][]{promptbox, title=System Prompt, #1}
\definecolor{nvidiagreen}{HTML}{76B900}
\hypersetup{
    colorlinks=true,
    citecolor=teal,
    linkcolor=nvidiagreen,  
    urlcolor=Green,
}

\AtBeginDocument{%
}

\newcommand{\appref}[1]{\hyperref[#1]{Appendix~\ref*{#1}}}
\newcommand{\dpos}[1]{\textit{\textcolor{deltateal}{$+#1$}}}
\newcommand{\dneg}[1]{\textit{\textcolor{newred}{$-#1$}}}
\newcommand{\dposavg}[1]{\textit{\textbf{\textcolor{deltateal}{$+#1$}}}}

\newcommand{\think}{\texttt{<think>}\xspace}
\newcommand{\thinkr}{\texttt{</think>}\xspace}
\newcommand{\toolcall}{\texttt{<tool\_call>}\xspace}
\newcommand{\toolcallr}{\texttt{</tool\_call>}\xspace}
\newcommand{\tool}{\texttt{<tool>}\xspace}
\newcommand{\toolr}{\texttt{</tool>}\xspace}
\newcommand{\answer}{\texttt{<answer>}\xspace}
\newcommand{\answerr}{\texttt{</answer>}\xspace}

\newcommand{\gap}{Thinking-Acting Gap\xspace}

\newcommand{\toolresample}{tool-call resampling\xspace}
\newcommand{\Toolresample}{Tool-call resampling\xspace}

\usepackage{amsmath,amsfonts,bm}

\def\eqref#1{equation~\ref{#1}}

\def\1{\bm{1}}

\def\va{{\bm{a}}}

\def\vo{{\bm{o}}}

\def\vt{{\bm{t}}}

\def\vx{{\bm{x}}}
\def\vy{{\bm{y}}}

\DeclareMathAlphabet{\mathsfit}{\encodingdefault}{\sfdefault}{m}{sl}
\SetMathAlphabet{\mathsfit}{bold}{\encodingdefault}{\sfdefault}{bx}{n}

\newif\ifmathvr
\mathvrtrue

\ifmathvr
  \newcommand{\dPassFourTwoB}{$+2.8$}
  \newcommand{\dPassFourFourB}{$+2.3$}
  \newcommand{\dPassFourEightB}{$+1.8$}
  \newcommand{\dPassOneTwoB}{$+1.1$}
  \newcommand{\dPassOneFourB}{$+1.4$}
  \newcommand{\dPassOneEightB}{$+1.8$}
  \newcommand{\eightBvsAgentPassFour}{$75.8$ vs.\ $75.1$}
  \newcommand{\eightBPassOneReachPct}{$99\%$}
  \newcommand{\dReasoningEightB}{$+0.6$}
  \newcommand{\dPerceptionEightB}{$+2.9$}
  \newcommand{\dSearchEightB}{$+1.3$}
  \newcommand{\dAgentTwoB}{$-1.3$ / $+1.1$}
  \newcommand{\dAgentFourB}{$+5.2$ / $+9.2$}
  \newcommand{\dAgentEightB}{$+4.7$ / $+9.8$}
  \newcommand{\dAgentThirtyTwoB}{$+6.7$ / $+10.1$}
  \newcommand{\dSFTinitPassOne}{$+5.3$ at 2B, $+5.2$ at 4B, $+1.3$ at 8B}
  \newcommand{\dSFTinitPassFour}{$+6.8$, $+4.9$, $+2.4$}
  \newcommand{\dRLAfterSFTPassOne}{$+4.5$ at 2B, $+1.2$ at 4B, $+2.0$ at 8B}
  \newcommand{\dRLAfterSFTPassFour}{$+2.3$, $+0.0$, $+1.8$}
  \newcommand{\toolUtilAxpoAvg}{$81.9$ / $86.9$ / $82.9$}
  \newcommand{\toolUtilHardReasoning}{MathVision, DynaMath, Math-VR}
  \newcommand{\toolUtilHardReasoningGap}{$+30/+33/+22$}
\else
  \newcommand{\dPassFourTwoB}{$+2.5$}
  \newcommand{\dPassFourFourB}{$+2.6$}
  \newcommand{\dPassFourEightB}{$+1.8$}
  \newcommand{\dPassOneTwoB}{$+0.7$}
  \newcommand{\dPassOneFourB}{$+1.4$}
  \newcommand{\dPassOneEightB}{$+1.9$}
  \newcommand{\eightBvsAgentPassFour}{$75.9$ vs.\ $75.4$}
  \newcommand{\eightBPassOneReachPct}{$98\%$}
  \newcommand{\dReasoningEightB}{$+1.0$}
  \newcommand{\dPerceptionEightB}{$+2.9$}
  \newcommand{\dSearchEightB}{$+1.3$}
  \newcommand{\dAgentTwoB}{$-1.4$ / $+1.5$}
  \newcommand{\dAgentFourB}{$+4.7$ / $+8.9$}
  \newcommand{\dAgentEightB}{$+3.4$ / $+8.8$}
  \newcommand{\dAgentThirtyTwoB}{$+6.6$ / $+10.1$}
  \newcommand{\dSFTinitPassOne}{$+5.0$ at 2B, $+5.2$ at 4B, $+0.9$ at 8B}
  \newcommand{\dSFTinitPassFour}{$+6.5$, $+5.0$, $+2.0$}
  \newcommand{\dRLAfterSFTPassOne}{$+4.4$ at 2B, $+1.2$ at 4B, $+1.8$ at 8B}
  \newcommand{\dRLAfterSFTPassFour}{$+2.0$, $-0.1$, $+1.6$}
  \newcommand{\toolUtilAxpoAvg}{$84.8$ / $95.7$ / $88.4$}
  \newcommand{\toolUtilHardReasoning}{MMMU-Pro, MathVision, DynaMath}
  \newcommand{\toolUtilHardReasoningGap}{$+30/+30/+33$}
\fi

\title{Agent Explorative Policy Optimization for Multimodal Agentic Reasoning}

\author{%
  Minki Kang$^{1,2*}$,
  Shizhe Diao$^{1}$,
  Ryo Hachiuma$^{1}$,
  Sung Ju Hwang$^{2}$,
  Pavlo Molchanov$^{1}$,
  Yu-Chiang Frank Wang$^{1}$,
  Byung-Kwan Lee$^{1\dagger}$\\
  $^1$NVIDIA \quad $^2$KAIST
}
\correspondingauthor{$^*$Work done during internship. $^\dagger$Project Lead.\\
Correspondence to <\texttt{minkikang@kaist.ac.kr}, \texttt{byungkwanl@nvidia.com}>}

\begin{abstract}
\textbf{Abstract}\\
Vision-language models with extended reasoning succeed on complex problems, but many real-world problems require external tools that internal reasoning alone often cannot resolve. Agentic reasoning therefore interleaves two behaviors with a structural asymmetry: thinking (the self-contained default) and tool use (a high-variance auxiliary acting). We refer to this asymmetry as the \emph{Thinking-Acting Gap}. Under standard RL recipes like GRPO, the gap manifests as two diagnostic symptoms during training: tool use is attempted on only $\sim$$30\%$ of rollouts, and when attempted, the tool-using rollouts within a group are all-wrong on $\sim$$40\%$ of questions, suppressing the learning signal at the tool calls that needed it. We propose \textbf{AXPO} (\textbf{A}gent e\textbf{X}plorative \textbf{P}olicy \textbf{O}ptimization): for each all-wrong tool-using subgroup, AXPO fixes the thinking prefix and resamples the tool call and its continuation, paired with uncertainty-based prefix selection.
Across nine multimodal benchmarks and three scales of Qwen3-VL-Thinking, SFT\,+\,AXPO outperforms SFT\,+\,GRPO at average (\dPassOneEightB\,pp Pass@1 and \dPassFourEightB\,pp Pass@4 at 8B on average) and 8B with SFT\,+\,AXPO surpasses the 32B Base on Pass@4 with $4\times$ fewer parameters.
The project page is available at \href{https://byungkwanlee.github.io/AXPO-page/}{link}.
\end{abstract}

\begin{document}

\maketitle

\begin{figure}[h]
    \centering
    \includegraphics[width=0.99\linewidth]{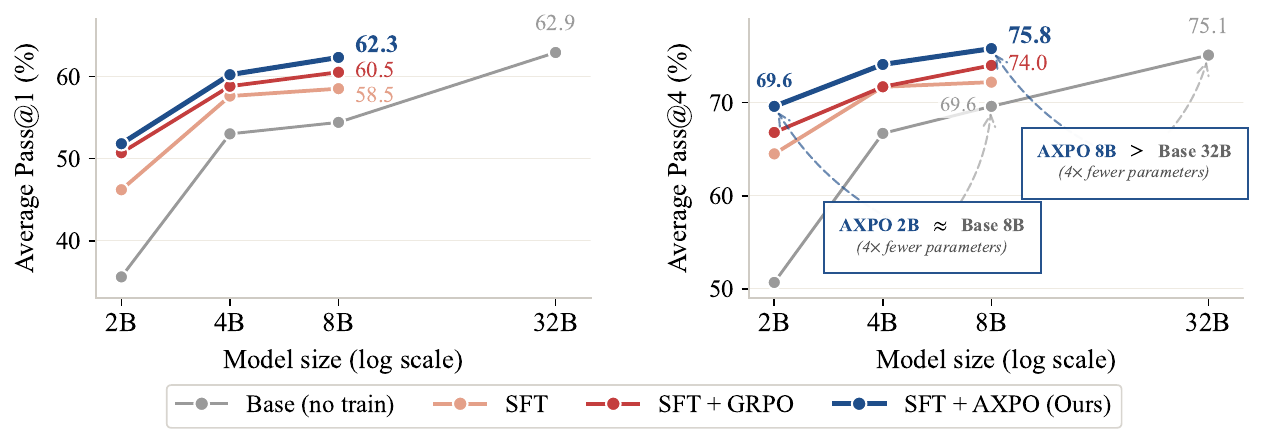}
    \vspace{-0.1in}
    \caption{\textbf{Main result: AXPO narrows the agentic gap to a 4$\times$ larger baseline.} Average Pass@1 (left) and Pass@4 (right) over nine multimodal benchmarks across Qwen3-VL-Thinking sizes. \emph{At 8B, AXPO surpasses the 32B Base baseline on Pass@4} while outperforming GRPO at every scale.}
    \label{fig:main}
    \vspace{0in}
\end{figure}

\section{Introduction}
\label{sec:intro}

Large language models (LLMs) and vision-language models (VLMs) with extended reasoning capabilities have made remarkable progress, producing chains of thought that rival much larger models on mathematical and scientific reasoning benchmarks~\citep{r1, o1, s1}.
These reasoning models invest test-time compute into internal search and reflection to arrive at accurate answers.
However, many real-world problems lie beyond what internal reasoning alone can resolve: retrieving up-to-date knowledge, performing complex numerical computation, and understanding fine-grained visual details all require access to external resources~\citep{Search-R1, Search-o1, PoT, mini-o3, vstar}.
Tools such as web search, code interpreters, and image manipulation utilities can fill this gap, enabling what we call \textbf{\emph{Multimodal Agentic Reasoning}}: the model autonomously decides when to think further and when to act through tools, much as it learns to allocate thinking budget, but now extended to interaction with environments~\citep{AgenticReasoning, tora, TIR, pyvision, toolorchestra, visualsketchpad}.

\begin{figure}[t]
    \centering
    \includegraphics[width=0.99\linewidth]{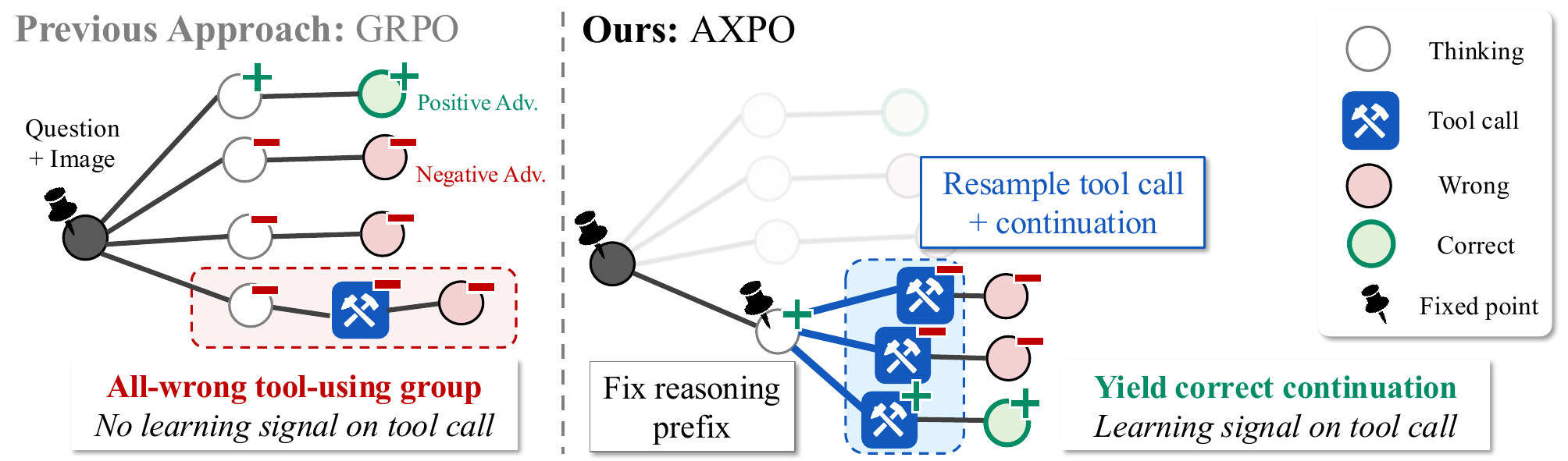}
    \vspace{-0.in}
    \caption{\textbf{Concept.} AXPO concentrates exploration on the tool call for agentic reasoning. \textbf{Left:} under GRPO, tool-using rollouts are scarce and often fail, leaving no positive learning signal at the tool call where it is most needed. \textbf{Right:} AXPO fixes the thinking prefix from the failed rollout and resamples the tool call and its continuation, recovering a correct rollout and producing a positive learning signal on the tool call.}
    \vspace{-0.26in}
    \label{fig:concept}
\end{figure}

Agentic reasoning carries a deeper asymmetry between its two constituent behaviors: \emph{Thinking} and \emph{Tool use (Acting)}.
\emph{Thinking} is the default mode, self-contained within the model's token space: every token is under its control and errors can be corrected by exploring a different chain of thought~\citep{R1-thoughtology,ahamoment,understandingreasoning}.
\emph{Tool use}, by contrast, is an auxiliary behavior that is simultaneously powerful but fragile: a short token sequence triggers an external response the model cannot predict or control, and small changes in the call can produce drastically different outcomes (e.g., invalid python code with wrong indent)~\citep{pyvision-rl}.
Reasoning models are heavily post-trained for thinking, but far less for acting.
We refer to this asymmetry between a safe default and a high-variance auxiliary as the \textbf{\emph{\gap}}. 

Due to this gap, eliciting agentic reasoning remains a challenge despite the  capabilities of recent reasoning VLMs~\citep{Qwen3VL}. Relying solely on prompting is often insufficient to ensure an appropriate tool call when needed. This necessitates post-training pipelines, such as Supervised Fine-Tuning (SFT) and Reinforcement Learning (RL), as evidenced in~\autoref{fig:main}.
SFT reliably teaches the \emph{mechanics} of tool use but struggles to teach the model to utilize tools better for situations the teacher's demonstrations did not cover. The resulting agent inherits the teacher's failure modes and often plateaus near its performance~\citep{AgentDistill, dagger}.
Therefore, the model requires a training signal that goes beyond imitation, which motivates utilizing RL with verifiable outcome rewards after SFT~\citep{pyvision-rl, AgenticReasoningRL, deepeyesv2}.

However, RL methods such as GRPO~\citep{GRPO,DAPO,prorl} are suboptimal to make models proficient at acting, because two patterns suppress the tool-call learning signal.
First, tool use remains a minority of rollouts compared with thinking-only (\autoref{fig:analysis-a}).
Second, when tool use is attempted, the entire group of rollouts is far more likely to fail together (\autoref{fig:analysis-b}).
Under GRPO, this asymmetry weakens the tool-call learning signal in both regimes: tool-using rollouts trail thinking-only successes in mixed groups (negative advantage) and zero out in all-wrong groups~\citep{pope, sage}.
These observations motivate targeted exploration for tool use during RL. This raises a critical question: \emph{how can we effectively induce such exploration?}
The \gap suggests the answer.
Thinking is the native mode of the model and the tool call is the separate mode that turns an intent into a concrete action where the same intent admits diverse implementations (\autoref{fig:analysis-c}).

Building on this observation, our key insight is that resampling at the tool call lets the model explore diverse agentic reasoning: we anchor the thinking prefix and resample the tool call and its continuation (\autoref{fig:concept}).
We call this \emph{\toolresample}.
The operation has no analog in pure reasoning, where there is no comparable boundary between a committed intent and a high-variance action.
When a tool-using trajectory fails, the thinking prefix may be entirely sound. Re-deriving the full trajectory from scratch wastes budget on prefixes that were not the bottleneck, whereas \toolresample concentrates exploration precisely where the \gap manifests. We propose \textbf{AXPO} (\textbf{A}gent e\textbf{X}plorative \textbf{P}olicy \textbf{O}ptimization) based on this principle to narrow the gap during RL.

Concretely, AXPO instantiates three design choices on top of a small extra rollout budget: (1) \emph{where to resample}, only on all-wrong tool-using subgroups where a single recovery flips the GRPO advantage;
(2) \emph{which prefix}, the highest-uncertainty ones; and (3) \emph{how resamples enter the gradient}, a per-prefix GRPO advantage on continuations and a binary recovery indicator on the source prefix.
Unlike prior reward-shaping methods~\citep{pyvision-rl, deepeyesv2, otc} or rollout scaling methods~\citep{brorl, arpo}, AXPO restructures the rollout distribution itself at the locus of the \gap.

Across nine multimodal benchmarks at Qwen3-VL-Thinking 2B/4B/8B~\citep{Qwen3VL}, SFT\,+\,AXPO outperforms SFT\,+\,GRPO at average with Pass@4 gains of \dPassFourTwoB/\dPassFourFourB/\dPassFourEightB\,pp.
\textbf{At 8B, SFT\,+\,AXPO surpasses the 32B Base} on Pass@4 and reaches \eightBPassOneReachPct{} of its Pass@1, despite using $4\times$ fewer parameters (\autoref{fig:main}). The improvement concentrates on tool-intensive benchmarks, while preserving the performance improvement in reasoning benchmarks compared to SFT\,+\,GRPO baseline. Notably, $+25\%$ extra resampling budget with AXPO delivers $+1.1$\,pp over $+100\%$ extra rollout budget with GRPO, confirming that the gain comes from where compute is spent.

Our contributions are as follows:
\begin{itemize}[leftmargin=*, itemsep=2pt]
    \item We identify the \textbf{\gap}, an asymmetry between thinking and tool use that manifests in group-based agentic RL as two symptoms: tool use is attempted on 30\% of rollouts and tool-using subgroups (the tool-using rollouts within a group) are all-wrong on 40\% of questions, yielding non-positive advantage on tool-call tokens.
    \item We propose \textbf{AXPO}, an RL algorithm that closes the gap via \emph{\toolresample}: for all-wrong tool-using subgroups, AXPO fixes the thinking prefix at uncertain tool calls and resamples from them, provably dominating from-scratch sampling on tool-use coverage at fixed budget.
    \item We validate AXPO on nine multimodal benchmarks across three scales of Qwen3-VL-Thinking, with gains for all sizes compared to GRPO baseline, and the \textbf{resulting 2B model matches the 8B Base and the 8B model surpasses the 32B Base on Pass@4}.
\end{itemize}

\section{Analysis of RL in Agentic Reasoning}
\subsection{Preliminaries}

\paragraph{Agentic Reasoning.}
We study the agentic reasoning of Vision-Language Models (VLMs) $\pi_\theta$ given an input $\vx$ (text question with visual image) and instruction prompt (\ref{app:tools}), interleaving thinking and tool use. At turn $t$, the model emits a thinking segment $\vt_t$ (in \think$\,\ldots\,$\thinkr) and an action $\va_t$ (a tool call formatted as JSON with tool type and arguments, in \toolcall$\,\ldots\,$\toolcallr), then receives an observation $\vo_t = \texttt{exec}(\va_t)$ (in \tool$\,\ldots\,$\toolr). 
We write $\vy_t = \vt_t \va_t \vo_t$ with $\vt_t \va_t \sim \pi_\theta(\cdot \mid \vx, \vy_{<t})$. This continues until the final answer (in \answer$\,\ldots\,$\answerr) is emitted or the turn limit is reached, yielding $\tau = (\vx, \vy_1, \ldots, \vy_T, \vy_{\text{ans}})$. Our tool set covers three standard~\citep{mini-o3, pyvision-rl, deepeyesv2}: a Python interpreter, a web search, and an image zoom-in tool.

\paragraph{RL training with GRPO.}
\label{sec:prelim-rl}

Our starting point is the standard agentic post-training pipeline~\citep{pyvision-rl, deepeyesv2}: supervised fine-tuning (SFT) on expert trajectories followed by reinforcement learning (RL) with verifiable rewards. Given an input $\vx$, the policy $\pi_\theta$ generates a group of $N$ rollouts $\{\tau_i\}_{i=1}^N$, each receiving a binary outcome reward $r_i \in \{0, 1\}$ based on answer correctness. Group Relative Policy Optimization (GRPO)~\citep{GRPO} computes a group-normalized advantage $A_i = (r_i - \mathrm{mean}(\{r_j\}_{j=1}^N)) / \mathrm{std}(\{r_j\}_{j=1}^N),$
which is assigned uniformly to every token in $\tau_i$ and used to update the policy via the PPO-clip surrogate $\mathcal{L}_{\text{clip}}$~\citep{PPO}. We defer the explicit form of $\mathcal{L}_{\text{clip}}$ in \appref{app:training-setup}.

\subsection{Diagnosing the \gap}
\label{sec:gap}
\begin{figure}[t]
    \centering
    \begin{subfigure}[t]{0.32\linewidth}
        \centering
        \includegraphics[width=\linewidth]{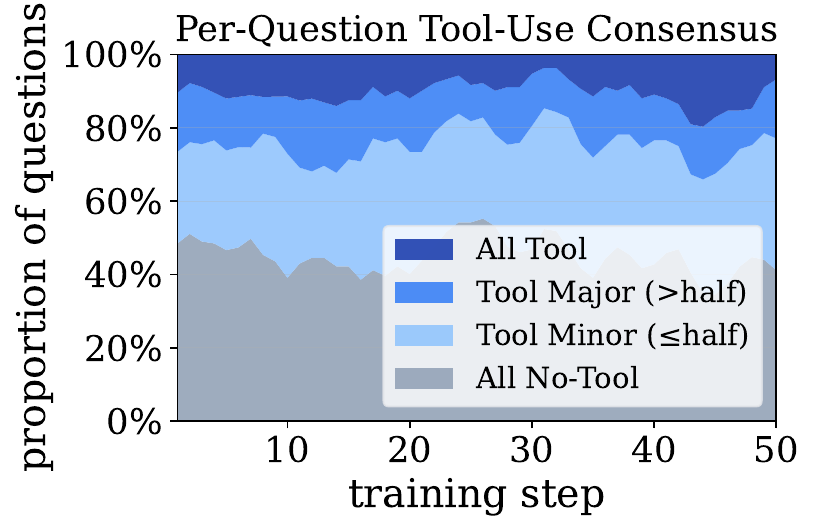}
        \vspace{-0.25in}
        \caption{Proportion of questions by tool-using rollout count (8 rollouts/question) across RL steps under GRPO.}
        \label{fig:analysis-a}
    \end{subfigure}
    \hfill
    \begin{subfigure}[t]{0.32\linewidth}
        \centering
        \includegraphics[width=\linewidth]{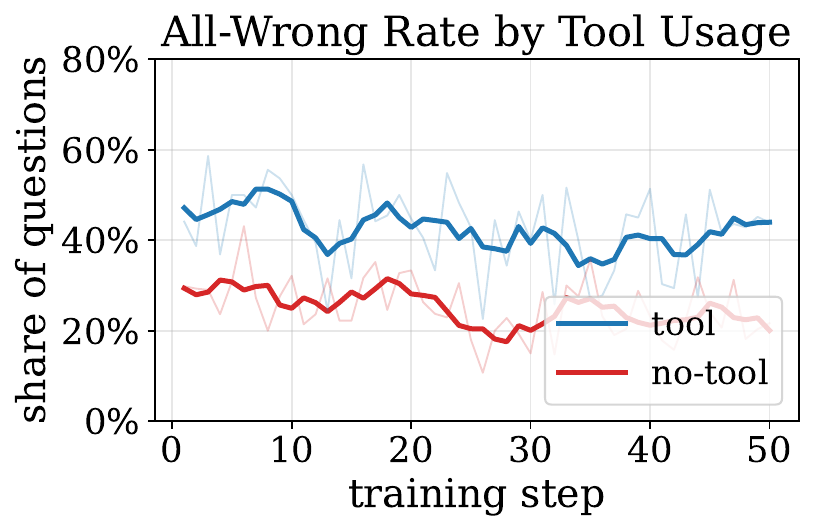}
        \vspace{-0.25in}
        \caption{All-wrong rate by subgroup type under GRPO: the tool-using fails together far more than the no-tool.}
        \label{fig:analysis-b}
    \end{subfigure}
    \hfill
    \begin{subfigure}[t]{0.32\linewidth}
        \centering
        \centering
        \includegraphics[width=\linewidth]{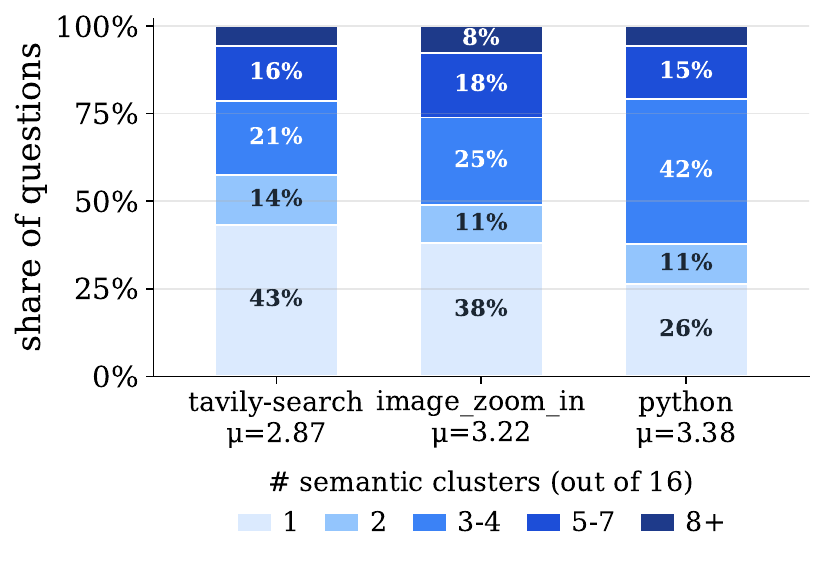}
        \vspace{-0.25in}
        \caption{Semantic cluster count of 16 resampled tool calls from a fixed failed-rollout prefix, per tool type.}
        \label{fig:analysis-c}
    \end{subfigure}
    \vspace{-0.07in}
    \caption{\textbf{Empirical analysis of diagnosing the \gap} (\autoref{sec:gap}). 
    \textbf{(a)} Tool use is under-attempted. \textbf{(b)} The tool-using subgroup (the tool-using rollouts within a GRPO group) fails all-together far more often than the no-tool subgroup. \textbf{(c)} Resampling at a fixed prefix preserves substantive diversity in the tool call, confirming it as a real divergence point.}
    \vspace{-0.1in}
    \label{fig:analysis}
\end{figure}

We now ground the \gap in three diagnostic measurements collected during RL training (\autoref{fig:analysis}). Our starting point is the SFT checkpoint, trained with GRPO~\citep{GRPO} on outcome rewards.

\paragraph{Tool use is under-attempted.}
A defining property of agentic reasoning, unlike conventional agentic RL~\citep{arpo, ragen2}, is that tool use is \emph{optional}: the model can either produce a \toolcall tag or skip straight to \answer. This optionality makes action fragile under RL.
We track the fraction of rollouts that contain at least one tool call over the course of RL. As shown in~\autoref{fig:analysis-a}, which bins each question by its tool-using rollout count among $N=8$, the corresponding rollout-level tool-use rate is roughly 27\% at step~1 and stays in the 20--35\% range over RL training, well below the thinking-only majority.
GRPO exerts no pressure to grow the tool-using minority, and the behavior that distinguishes agentic from non-agentic reasoning remains under-represented in each group.

\paragraph{When attempted, the tool-using subgroup is prone to being all-wrong.}
Within a GRPO group of $N$ rollouts for one question, we call the rollouts that emit a tool call the \emph{tool-using subgroup} and the rest the \emph{no-tool subgroup}, and the tool-using subgroup is \emph{all-wrong} when none of its rollouts is correct, even if some no-tool rollouts in the same group succeed. Conditional on a question where tool use is attempted, how often does the tool-using subgroup fail together? \autoref{fig:analysis-b} shows this all-wrong rate is $\sim$40\% for tool-using subgroups compared to $\sim$25\% for no-tool subgroups. Under group-normalized advantage, this asymmetry yields non-positive advantage on tool-call tokens in both regimes: in mixed groups they trail the no-tool success signal, and in fully all-wrong groups the numerator $r_i - \operatorname{mean}(\{r_j\})$ collapses to zero. Either way, effective learning signal on tool calls is suppressed on the questions where tool use would matter most.

\paragraph{Tool calls admit substantive resampling at a fixed thinking.}
Resampling at the tool call is only worthwhile if it is a real divergence point under the policy.
If the thinking prefix already commits to the concrete action, resampling at the tool-call boundary changes nothing.
We probe this by decomposing a trajectory at the first tool-call boundary: we fix its thinking prefix $\vt_1$, draw 16 continuations from $\pi_\theta(\cdot \mid \vx, \vt_1)$, and cluster the resulting tool calls into semantic classes by an LLM judge that groups calls by their semantic intent. \autoref{fig:analysis-c} shows that the resampled tool calls do not collapse to a single class. Across all tools, the 16 samples occupy 2.9--3.4 distinct semantic clusters on average. The thinking prefix therefore does not determine the exact tool call and the tool call is an under-explored divergence point whose exploration must be supplied externally.

\begin{takeawaybox}{Takeaways: GRPO under-trains the tool call}
\begin{itemize}[leftmargin=*, itemsep=1pt, topsep=2pt]
    \item Tool use remains a minority behavior under GRPO: $\sim$20--35\% of rollouts across RL steps, well below the thinking-only majority (\autoref{fig:analysis-a}).
    \item When attempted, the tool-using subgroup is all-wrong on $\sim$40\% of questions (vs.\ 25\% for the no-tool subgroup), so tool-call tokens get negative advantage in mixed groups and zero in fully all-wrong groups (\autoref{fig:analysis-b}).
    \item Resampling from a fixed prefix preserves substantive diversity in the tool call (2.9--3.4 semantic clusters per 16 samples in~\autoref{fig:analysis-c}): the tool call is a real divergence point.
\end{itemize}
\end{takeawaybox}

\section{Agent eXplorative Policy Optimization (AXPO)}
\label{sec:method}

\subsection{\Toolresample}
\label{sec:branching}

In agentic reasoning, the policy must learn both thinking and tool use where the latter is the under-trained behavior under GRPO (\autoref{sec:gap}). Rather than scaling rollouts $N$ to generate tool-using successes uniformly, we commit to a thinking prefix that promises a tool call and concentrate stochasticity on the continuation.

Concretely, let $\tau^{\text{src}}$ denote a \textbf{tool-using} rollout from the standard GRPO group of $\vx$, which we call the \emph{source rollout}, and let $\vt_1^{\text{src}}$ be its prefix that includes the opening \toolcall tag, so that decoding from $\vt_1^{\text{src}}$ is constrained to produce a valid tool call.
\emph{\Toolresample} draws $K$ continuations $\{\vy_k^{\text{res}}\}_{k=1}^{K} \sim \pi_\theta(\cdot \mid \vx, \vt_1^{\text{src}})$, executes the tool calls inside each, and rolls every resampled trajectory forward until an \answer tag or the turn limit. Each resampled trajectory $\tau_k^{\text{res}} = (\vt_1^{\text{src}}, \vy_k^{\text{res}})$ therefore shares the fixed thinking prefix with $\tau^{\text{src}}$ and diverges at the tool call, concentrating stochasticity on the action that the thinking was preparing for. Each resampled trajectory receives the same binary outcome reward $r_k^{\text{res}} \in \{0,1\}$ as a standard rollout.

\paragraph{Theoretical motivation.}
\textit{Why grow tool-using rollout coverage?} Recent analysis~\citep{brorl} shows that scaling per-prompt rollouts $N$ yields a policy-improvement guarantee that grows with the correct-token mass the group covers. 
However, our learning target is correct \emph{tool-using} rollouts carry positive gradient signal for tool-call tokens, so coverage at the tool call is bounded by the tool-use rate $q = \Pr_{\pi_\theta}[\texttt{tool use} \mid \vx]$ rather than $N$, and a fraction $1-q$ of any raw sampling budget is wasted on non-tool rollouts that cannot teach the tool call. \Toolresample sidesteps this waste by committing to a prefix $\vt_1^{\text{src}}$ that has already crossed the \toolcall boundary: every resampled continuation from $\vt_1^{\text{src}}$ is tool-using by construction, so the full resampling budget lands on the signal we actually want.

Let $p(\vt_1^{\text{src}}) = \Pr_{\pi_\theta}[R=1 \mid \vx, \vt_1^{\text{src}}]$ be the success probability of a continuation from prefix $\vt_1^{\text{src}}$, and $p^{\text{tool}} = \Pr_{\pi_\theta}[R=1 \mid \vx,\, \texttt{tool use}]$ the per-tool-using-rollout success rate of standard rollout. The following proposition makes the gain precise.

\begin{proposition}[Tool-call resampling dominates raw sampling at recovering tool-using successes]
\label{prop:branching}
Let $\vt_1^{\text{src}}$ be a prefix that has crossed the tool call boundary with $p(\vt_1^{\text{src}}) \geq q\, p^{\text{tool}}$. Then for any $N \geq 1$,
\vspace{-0.05in}
\begin{equation}
    \underbrace{1 - \bigl(1 - p(\vt_1^{\text{src}})\bigr)^N}_{\text{resampling: $N$ tool-using continuations}}
    \;\geq\;
    \underbrace{1 - (1 - q\, p^{\text{tool}})^N}_{\text{raw: $N$ rollouts, $q$-fraction tool-using}},
\end{equation}
with strict inequality whenever $p(\vt_1^{\text{src}}) > q\, p^{\text{tool}}$ and $q\, p^{\text{tool}} \in (0,1)$. Both sides measure the probability that the group contains at least one correct tool-using rollout.
\end{proposition}

The key mechanism is $q \to 1$ by construction: every resampled continuation from $\vt_1^{\text{src}}$ is tool-using, so $(1-q)$ waste factor vanishes and the threshold $q\, p^{\text{tool}}$ sits strictly below the average per-tool-using-rollout success $p^{\text{tool}}$, admitting prefixes at or below the conditional mean. Under GRPO, the left-hand side equals the probability that the tool-using subgroup produces a positive gradient, so resampling strictly raises the probability of recovering an all-wrong tool-using subgroup identified in \autoref{sec:gap} relative to raw sampling at any fixed $N$. Detailed analysis is in \appref{app:coverage}.

\subsection{Resample candidate selection}
\label{sec:selection}

\paragraph{Trigger and budget allocation.}
AXPO resamples only on groups in which the tool-using subgroup is non-empty and entirely wrong (an \emph{all-wrong tool-using subgroup}; no-tool rollouts in the same group may succeed). These are exactly the groups where tool-call tokens receive non-positive advantage under GRPO (\autoref{sec:gap}): zero when the entire group is all-wrong, negative when only no-tool rollouts succeed. A recovered continuation therefore delivers the largest gradient lift per resampled continuation. \autoref{prop:branching} would in principle admit groups with at least one correct tool-using rollout as well, but those already produce positive advantage on tool-call tokens and gain comparatively little from a marginal recovery. Resampling every triggered prefix would add up to $BNK$ continuations on top of $BN$ standard rollouts (batch size $B$), so AXPO caps the extra cost up to $r \cdot BN$ per step with $r < 1$ and allocates breadth-first: every triggered question receives its top-ranked prefix before any receives a second.

\paragraph{Uncertainty-based prefix ranking.}
Not every admitted prefix is worth resampling. \autoref{fig:analysis-c} shows $\sim$30\% of prefixes collapse to a single cluster across 16 resamples. AXPO therefore ranks candidates by the mean policy probability over the tool-call tokens in $\tau^{\text{src}}$ (a tractable uncertainty proxy~\citep{deepconf}) and resamples the lowest-confidence prefixes first (see \appref{app:conf-vs-entropy} for the choice of confidence over predictive entropy).

\subsection{Advantage calculation}
\label{sec:credit}

Given the resampled trajectories from \autoref{sec:selection}, we now specify how their advantages enter the gradient. Because each $\tau_k^{\text{res}}$ shares its prefix with the source rollout $\tau^{\text{src}}$, training both as independent rollouts would assign contradictory advantages to the prefix tokens: the failed source penalizes the prefix while a successful resample rewards it. AXPO decomposes the advantage streams so that every token is updated from exactly one source.

For each selected prefix $\vt_1^{\text{src}}$, the $K$ resampled continuations form an independent advantage group. We compute a per-prefix GRPO advantage
\begin{equation}
\label{eq:axpo-adv}
\hat{A}_k^{\text{res}}(\vt_1^{\text{src}}) \;=\; \frac{r_k^{\text{res}} - \mathrm{mean}\bigl(\{r_j^{\text{res}}\}_{j=1}^K\bigr)}{\mathrm{std}\bigl(\{r_j^{\text{res}}\}_{j=1}^K\bigr)}, \qquad k = 1, \ldots, K,
\end{equation}
which is applied only to the continuation tokens $\vy_k^{\text{res}}$ where the prefix thinking tokens are masked. The source trajectory's prefix tokens are updated through $\tau^{\text{src}}$ with a binary recovery reward that fires whenever resampling recovers at least one correct continuation:
\begin{equation}
\label{eq:axpo-recovery}
r^{\text{prefix}}(\vt_1^{\text{src}}) \;=\; \mathds{1}\!\left[\exists\, k \in \{1,\ldots,K\}:\; r_k^{\text{res}} = 1\right].
\end{equation}
This recovery reward replaces the original (zero) source-rollout reward in the source group's GRPO normalization, yielding a per-prefix advantage that we apply to the source's prefix tokens as follows:
\begin{equation}
\label{eq:axpo-prefix-adv}
\hat{A}^{\text{prefix}}(\vt_1^{\text{src}}) \;=\; \frac{r^{\text{prefix}}(\vt_1^{\text{src}}) - \mathrm{mean}\bigl(\{\tilde r_j\}_{j=1}^N\bigr)}{\mathrm{std}\bigl(\{\tilde r_j\}_{j=1}^N\bigr)},
\qquad
\tilde r_j \;=\; \begin{cases} r^{\text{prefix}}(\vt_1^{\text{src}}), & j = \text{source index}, \\ r_j, & \text{otherwise}, \end{cases}
\end{equation}
and the source continuation is masked. Combining both streams with the standard clipped surrogate from PPO~\citep{PPO} yields the resampling contribution for a selected prefix $\vt_1^{\text{src}}$ to maximize:
\begin{equation}
\label{eq:axpo-loss}
\mathcal{L}_{\text{AXPO}}(\vt_1^{\text{src}}) \;=\; \underbrace{\mathcal{L}_{\text{clip}}\bigl(\vt_1^{\text{src}};\, \hat{A}^{\text{prefix}}(\vt_1^{\text{src}})\bigr)}_{\text{prefix, source only}} \;+\; \sum_{k=1}^{K}\underbrace{\mathcal{L}_{\text{clip}}\bigl(\vy_k^{\text{res}};\, \hat{A}_k^{\text{res}}(\vt_1^{\text{src}})\bigr)}_{\text{resampled continuation}}.
\end{equation}
Non-triggered groups and rollouts not selected for resampling retain the standard GRPO loss from \autoref{sec:prelim-rl}.
\autoref{eq:axpo-loss} is added once per prefix selected in \autoref{sec:selection}.
The recovery-indicator $r^{\text{prefix}}$ avoids the gradient conflict of propagating mixed-sign continuation advantages back to the shared prefix: $\hat{A}^{\text{prefix}}$ is monotone in $r^{\text{prefix}}$, so the prefix is credited positively whenever at least one continuation succeeds, converting the coverage gain into gradient signal on the prefix tokens.

\section{Experiments}
\label{sec:experiments}

\subsection{Experimental setup}
\label{sec:setup}

\paragraph{Base models and tools.}
We use Qwen3-VL-Thinking~\citep{Qwen3VL} at three sizes (2B, 4B, 8B) as the base model across all methods. The tool set is the three families introduced in \autoref{sec:prelim-rl}: a Python interpreter, a web search engine (Tavily search API\footnote{https://www.tavily.com/}), and an image zoom-in tool. 
All tools share a unified \texttt{<tool\_call>} interface. Full prompt templates and tool schemas are in \appref{app:tools}.

\paragraph{Training data \& pipeline.}
SFT uses 64K correct trajectories drawn from ViRL~\citep{virl}, fvqa~\citep{fvqa}, and PyVision-RL~\citep{pyvision-rl}, generated by Qwen3-VL-32B-Thinking~\citep{Qwen3VL} where $\sim$25\% contain a tool call. RL uses a 16K filtered subset plus 22K questions from MMFineReason-hard~\citep{mmfinereason}.
We follow the SFT then RL pipeline of \autoref{sec:prelim-rl}. AXPO augments the rollout loop with \toolresample, holding all other hyperparameters constant against SFT\,+\,GRPO. Full details are in \appref{app:training-setup}.

\paragraph{Benchmarks and metrics.}
We evaluate on nine multimodal benchmarks grouped into three task categories: \textbf{Reasoning} (MathVision~\citep{mathvision}, DynaMath~\citep{dynamath}, Math-VR~\citep{mathvr}), \textbf{Perception} (V$^{\star}$~\citep{vstar}, VisualProbe~\citep{mini-o3}, HR-Bench-4K and HR-Bench-8K~\citep{HRBench}), and \textbf{Search} (HR-MMSearch~\citep{hrmmsearch}, MMSearch~\citep{mmsearch}); per-benchmark descriptions in \appref{app:benchmarks}. All benchmarks are evaluated in the agentic setting with the same tool interface used in training. We report Pass@1 and Pass@4 (\%), averaged over four rollouts per question, with answer matching logic tailored for each benchmark.

\paragraph{Baselines.}
We compare AXPO against four baselines: (1) \textbf{Base}: the base model only with prompting without additional training (2) \textbf{SFT}: the SFT checkpoint without RL (3) \textbf{GRPO}: RL from the base model without an SFT stage and (4) \textbf{SFT\,+\,GRPO}: the standard SFT-then-RL recipe.
Our method, \textbf{SFT\,+\,AXPO}: applies \toolresample and advantage calculation from~\autoref{sec:method} on top of the same SFT-then-RL pipeline.

\subsection{Main Results: AXPO narrows the agentic gap to a 4$\times$ larger baseline.}
\label{sec:main-results}
\begin{table*}[t]
    \caption{Main results --- Pass@1 (\%, $\uparrow$) on \textbf{nine multimodal benchmarks}, grouped into three categories: Reasoning, Perception, and Search. All methods use \textbf{Qwen3-VL-Thinking}~\citep{Qwen3VL} at four sizes (2B, 4B, 8B, 32B). Best result per column within each size block is \textbf{bolded}; second-best is \underline{underlined}. \textit{$\Delta$} rows show AXPO's gain over SFT\,+\,GRPO, the strongest comparable training recipe where AXPO outperforms it in most of evaluations. Pass@4 results are reported in \autoref{tab:main-p4}.}
    \vspace{-0.1in}
    \label{tab:main-p1}
    \centering
    \small
    \resizebox{\textwidth}{!}{
    \begin{tabular}{l ccc cccc cc >{\columncolor{mygray}}c}
        \toprule
        \multicolumn{1}{c}{} &
        \multicolumn{3}{c}{\textbf{Reasoning}} &
        \multicolumn{4}{c}{\textbf{Perception}} &
        \multicolumn{2}{c}{\textbf{Search}} &
        \multicolumn{1}{c}{\multirow{2}{*}[-3pt]{\textbf{Avg.}}} \\
        \cmidrule(lr){2-4} \cmidrule(lr){5-8} \cmidrule(lr){9-10}
        \textbf{Method} &
        \shortstack{Math\\Vision} & \shortstack{Dyna\\Math} & \shortstack{Math\\-VR} &
        V$^{\star}$ & \shortstack{Visual\\Probe} & \shortstack{HRBen.\\-4K} & \shortstack{HRBen.\\-8K} &
        \shortstack{HR-MM\\Search} & \shortstack{MM\\Search} & \multicolumn{1}{c}{} \\
        \midrule

        \multicolumn{11}{l}{\textit{Qwen3-VL-2B-Thinking}} \\
        \midrule
        Base
            & 18.6 & 60.5 & 16.9 & 63.0 & 5.0 & 64.6 & 55.9 & 8.2 & 27.7 & 35.6 \\
        GRPO
            & 33.7 & 64.9 & 34.6 & 67.1 & 23.8 & 65.0 & 59.4 & \textbf{18.4} & \textbf{41.4} & 45.4 \\
        SFT
            & 28.9 & 64.1 & 30.2 & 75.9 & 24.5 & 73.1 & 68.8 & 14.5 & 35.9 & 46.2 \\
        \quad+\,GRPO
            & \underline{34.8} & \underline{66.4} & \underline{35.8} & \textbf{81.7} & \textbf{36.1} & \underline{74.3} & \underline{70.6} & 15.7 & \underline{40.6} & \underline{50.7} \\
        \rowcolor{myblue}
        \quad+\,\textbf{AXPO}\,\scalebox{0.8}{\textit{(Ours)}}
            & \textbf{35.0} & \textbf{66.9} & \textbf{38.9} & \underline{81.3} & \underline{35.1} & \textbf{77.9} & \textbf{72.4} & \underline{18.1} & 40.3 & \textbf{51.8} \\
        \rowcolor{myblue!50}
        \textit{\;\;$\Delta$ vs.\ SFT\,+\,GRPO}
            & \dpos{0.2} & \dpos{0.5} & \dpos{3.1} & \dneg{0.4} & \dneg{1.0} & \dpos{3.6} & \dpos{1.8} & \dpos{2.4} & \dneg{0.3} & \dposavg{1.1} \\
        \midrule

        \multicolumn{11}{l}{\textit{Qwen3-VL-4B-Thinking}} \\
        \midrule
        Base
            & 48.0 & \underline{76.6} & 50.8 & 80.6 & 24.8 & 73.6 & 67.5 & 16.9 & 37.8 & 53.0 \\
        GRPO
            & \underline{52.5} & 76.2 & \textbf{57.9} & 82.7 & 29.7 & 75.5 & 71.1 & 12.7 & 23.8 & 53.6 \\
        SFT
            & 49.1 & 76.3 & 55.3 & 87.0 & 34.7 & \underline{79.5} & \underline{74.1} & \textbf{20.6} & 41.5 & 57.6 \\
        \quad+\,GRPO
            & 51.6 & \textbf{77.2} & \underline{57.4} & \textbf{90.2} & \underline{38.0} & 78.8 & 73.8 & 20.0 & \underline{42.3} & \underline{58.8} \\
        \rowcolor{myblue}
        \quad+\,\textbf{AXPO}\,\scalebox{0.8}{\textit{(Ours)}}
            & \textbf{52.8} & 76.4 & 57.2 & \underline{88.9} & \textbf{43.6} & \textbf{80.9} & \textbf{78.3} & \underline{20.1} & \textbf{43.3} & \textbf{60.2} \\
        \rowcolor{myblue!50}
        \textit{\;\;$\Delta$ vs.\ SFT\,+\,GRPO}
            & \dpos{1.2} & \dneg{0.8} & \dneg{0.2} & \dneg{1.3} & \dpos{5.6} & \dpos{2.1} & \dpos{4.5} & \dpos{0.1} & \dpos{1.0} & \dposavg{1.4} \\
        \midrule

        \multicolumn{11}{l}{\textit{Qwen3-VL-8B-Thinking}} \\
        \midrule
        Base
            & 47.1 & 75.9 & 54.9 & 77.7 & 31.8 & 72.8 & 66.1 & 21.0 & 42.7 & 54.4 \\
        GRPO
            & 54.4 & 77.2 & 58.9 & 85.7 & \underline{40.1} & 77.1 & 73.9 & 20.6 & \textbf{45.1} & 59.2 \\
        SFT
            & 50.9 & 76.8 & 58.4 & 84.8 & 38.4 & \underline{79.5} & 74.4 & 23.0 & 40.6 & 58.5 \\
        \quad+\,GRPO
            & \underline{55.3} & \underline{78.2} & \underline{60.4} & \underline{87.7} & \underline{40.1} & \underline{79.5} & \underline{74.9} & \underline{24.4} & 44.0 & \underline{60.5} \\
        \rowcolor{myblue}
        \quad+\,\textbf{AXPO}\,\scalebox{0.8}{\textit{(Ours)}}
            & \textbf{56.1} & \textbf{79.0} & \textbf{60.6} & \textbf{87.8} & \textbf{45.8} & \textbf{83.3} & \textbf{77.0} & \textbf{25.9} & \underline{45.0} & \textbf{62.3} \\
        \rowcolor{myblue!50}
        \textit{\;\;$\Delta$ vs.\ SFT\,+\,GRPO}
            & \dpos{0.8} & \dpos{0.8} & \dpos{0.2} & \dpos{0.1} & \dpos{5.7} & \dpos{3.8} & \dpos{2.1} & \dpos{1.5} & \dpos{1.0} & \dposavg{1.8} \\
        \midrule

        \multicolumn{11}{l}{\textit{Qwen3-VL-32B-Thinking} (inference-only baseline)} \\
        \midrule
        Base
            & 56.5 & 83.3 & 64.1 & 89.1 & 40.3 & 85.3 & 78.9 & 22.8 & 46.1 & 62.9 \\

        \bottomrule
    \end{tabular}
    }
\vspace{-0.15in}
\end{table*}

\autoref{tab:main-p1} reports Pass@1 across nine benchmarks at four model sizes (per-benchmark Pass@4 results in \autoref{tab:main-p4}, \appref{app:pass4-table}). SFT\,+\,AXPO outperforms SFT\,+\,GRPO at average performance on Pass@1 (\dPassOneTwoB{} / \dPassOneFourB{} / \dPassOneEightB\,pp at 2B/4B/8B). On Pass@4, SFT\,+\,AXPO also beats SFT\,+\,GRPO at every scale (\dPassFourTwoB{} / \dPassFourFourB{} / \dPassFourEightB\,pp). \textbf{At 8B, SFT\,+\,AXPO reaches \eightBPassOneReachPct{} of the 32B Base's Pass@1 and surpasses it on Pass@4 (\eightBvsAgentPassFour), despite having $4\times$ fewer parameters}. AXPO is the only recipe whose 8B point reaches the 32B line on Pass@1, and the only one whose Pass@4 crosses it (\autoref{fig:main}).

Decomposing the 8B Pass@1 column of \autoref{tab:main-p1}, SFT\,+\,AXPO gains \dReasoningEightB\,pp on Reasoning, \dPerceptionEightB\,pp on Perception, and \dSearchEightB\,pp on Search over the SFT\,+\,GRPO baseline.
In relative terms, the gain concentrates in Perception ($+4\%$) and is smallest on Reasoning ($+1\%$). Perception is where image-tool use most directly determines correctness, while on Reasoning the bottleneck is often the thinking rather than the tool call.
A detailed discussion of the results is provided in~\appref{app:decomp-recipe}.

\subsection{Ablations: All components of AXPO matter}
\label{sec:ablations}

\autoref{tab:ablation} ablates each AXPO design choice on 8B. Every component contributes: removing prefix fixing, the resampling target, uncertainty-based prefix selection, prefix credit, or the per-prefix advantage group all degrade both Pass@1 and Pass@4 average. Per-row analysis is in~\appref{app:ablation-details}.

\subsection{AXPO outperforms alternative RL recipes}
\label{sec:comparison}
\begin{table*}[t]
    \caption{\textbf{AXPO ablations} on \textbf{Qwen3-VL-8B-Thinking}. Each row removes (or alters) one AXPO design choice from the full method. All rows share the SFT initialization. Best per column is \textbf{bolded}; second-best \underline{underlined}.}
    \vspace{-0.1in}
    \label{tab:ablation}
    \centering
    \small
    \resizebox{\textwidth}{!}{
    \begin{tabular}{l cc cc cc cc >{\columncolor{mygray}}c>{\columncolor{mygray}}c}
        \toprule
        \textbf{Method} &
        \multicolumn{2}{c}{MathVision} &
        \multicolumn{2}{c}{V$^{\star}$} &
        \multicolumn{2}{c}{VisualProbe} &
        \multicolumn{2}{c}{HR-MMSearch} &
        \multicolumn{2}{c}{\textbf{Average}} \\
        \cmidrule(lr){2-3} \cmidrule(lr){4-5} \cmidrule(lr){6-7} \cmidrule(lr){8-9} \cmidrule(lr){10-11}
        SFT + & P@1 & P@4 & P@1 & P@4 & P@1 & P@4 & P@1 & P@4 & \multicolumn{1}{c}{P@1} & \multicolumn{1}{c}{P@4} \\
        \midrule

        AXPO\,w/o prefix fix (rollout from scratch)
            & 56.0 & 75.0 & 86.9 & \textbf{96.9} & 43.6 & \textbf{68.9} & \underline{21.9} & 35.7 & \underline{52.1} & \underline{69.1} \\
        \phantom{AXPO\,}w/\phantom{o} no-tool rollout resampling
            & 54.9 & 71.4 & \textbf{89.1} & \underline{96.3} & 41.5 & 62.3 & 21.1 & 33.4 & 51.7 & 65.9 \\
        \phantom{AXPO\,}w/o uncertainty rank
            & 53.0 & 72.4 & 85.7 & 93.2 & \underline{43.9} & 63.2 & 21.4 & 34.4 & 51.0 & 65.8 \\
        \phantom{AXPO\,}w/o prefix credit
            & \underline{57.2} & \textbf{76.6} & 86.3 & 95.3 & 40.6 & 62.3 & 21.6 & \underline{36.7} & 51.4 & 67.7 \\
        \phantom{AXPO\,}w/o separated advantage group
            & \textbf{57.7} & \textbf{76.6} & 87.2 & 95.8 & 40.1 & 61.3 & 19.8 & 34.4 & 51.2 & 67.0 \\
        \midrule
        \rowcolor{myblue}
        \textbf{AXPO}\,\scalebox{0.8}{\textit{(Ours)}}
            & 56.1 & \underline{75.7} & \underline{87.8} & 95.3 & \textbf{45.8} & \underline{67.9} & \textbf{25.9} & \textbf{42.0} & \textbf{53.9} & \textbf{70.2} \\

        \bottomrule
    \end{tabular}
    }
\end{table*}

\begin{table*}[t]
    \caption{\textbf{Comparison to alternative RL recipes} on \textbf{Qwen3-VL-8B-Thinking}. All rows share the SFT initialization. \textit{Reward shaping \& sampling} adds a per-tool-call reward or doubles the GRPO rollout budget. \textit{RL algorithms} replaces GRPO with another alternative. Best per column is \textbf{bolded}; second-best \underline{underlined}.}
    \vspace{-0.1in}
    \label{tab:comparison}
    \centering
    \small
    \resizebox{\textwidth}{!}{
    \begin{tabular}{l cc cc cc cc >{\columncolor{mygray}}c>{\columncolor{mygray}}c}
        \toprule
        \textbf{Method} &
        \multicolumn{2}{c}{MathVision} &
        \multicolumn{2}{c}{V$^{\star}$} &
        \multicolumn{2}{c}{VisualProbe} &
        \multicolumn{2}{c}{HR-MMSearch} &
        \multicolumn{2}{c}{\textbf{Average}} \\
        \cmidrule(lr){2-3} \cmidrule(lr){4-5} \cmidrule(lr){6-7} \cmidrule(lr){8-9} \cmidrule(lr){10-11}
        SFT + & P@1 & P@4 & P@1 & P@4 & P@1 & P@4 & P@1 & P@4 & \multicolumn{1}{c}{P@1} & \multicolumn{1}{c}{P@4} \\
        \midrule

        GRPO (baseline)
            & 55.3 & \underline{75.0} & 87.7 & \textbf{96.9} & 40.1 & 63.2 & \underline{24.4} & \underline{38.0} & 51.9 & \underline{68.3} \\
        \midrule
        \multicolumn{11}{l}{\textit{Reward shaping \& sampling}} \\
        GRPO\,+\,Tool penalty~\citep{otc}
            & 53.0 & 69.7 & 83.1 & 90.1 & 30.4 & 46.2 & 17.4 & 30.5 & 46.0 & 59.1 \\
        \phantom{GRPO\,}+\,Tool bonus~\citep{pyvision-rl, pixelreasoner, deepeyes}
            & 52.9 & 72.7 & 86.0 & 93.2 & 42.7 & 59.4 & 21.6 & 37.0 & 50.8 & 65.6 \\
        \phantom{GRPO\,}+\,rollout $2\times$ (\textbf{compute $2\times$})
            & \underline{55.6} & 73.4 & 87.0 & \underline{95.8} & \underline{43.6} & \underline{67.0} & 23.2 & 36.7 & \underline{52.4} & 68.2 \\
        \midrule
        \multicolumn{11}{l}{\textit{RL algorithms}} \\
        RLTF~\citep{rltf}
            & 42.4 & 65.5 & 79.8 & 84.3 & 30.2 & 42.5 & 20.6 & 31.8 & 43.3 & 56.0 \\
        CISPO~\citep{cispo}
            & 54.3 & 71.1 & 86.0 & 93.7 & 29.2 & 48.1 & 21.5 & 35.7 & 47.8 & 62.2 \\
        ARPO~\citep{arpo}
            & 53.2 & 71.7 & \textbf{89.3} & \underline{95.8} & 41.0 & 60.4 & 21.6 & 34.4 & 51.3 & 65.6 \\
        \midrule
        \rowcolor{myblue}
        \textbf{AXPO}\,\scalebox{0.8}{\textit{(Ours)}}
            & \textbf{56.1} & \textbf{75.7} & \underline{87.8} & 95.3 & \textbf{45.8} & \textbf{67.9} & \textbf{25.9} & \textbf{42.0} & \textbf{53.9} & \textbf{70.2} \\

        \bottomrule
    \end{tabular}
    }
    \vspace{-0.15in}
\end{table*}

\autoref{tab:comparison} compares AXPO against alternative training recipes: reward shaping (tool penalty / bonus), an extra rollout budget, and other RL algorithms (RLTF~\citep{rltf}, CISPO~\citep{cispo}, ARPO~\citep{arpo}). AXPO exceeds every alternative on both Pass@1 and Pass@4. The most direct more compute control, rollout $2\times$, doubles rollout budget yet underperforms than AXPO. The gain therefore comes from \emph{where} compute is spent, not from \emph{how much}. Reward-shaping methods both fall below the GRPO baseline of $51.9$, indicating the gap is not closed by reward shaping either in agentic reasoning. Detailed analysis is in~\appref{app:comparison-details}.

\subsection{AXPO narrows the \gap during RL}
\label{sec:dynamics}
\begin{figure}[t]
    \centering
    \begin{minipage}[b]{0.63\linewidth}
        \centering
        \includegraphics[width=\linewidth]{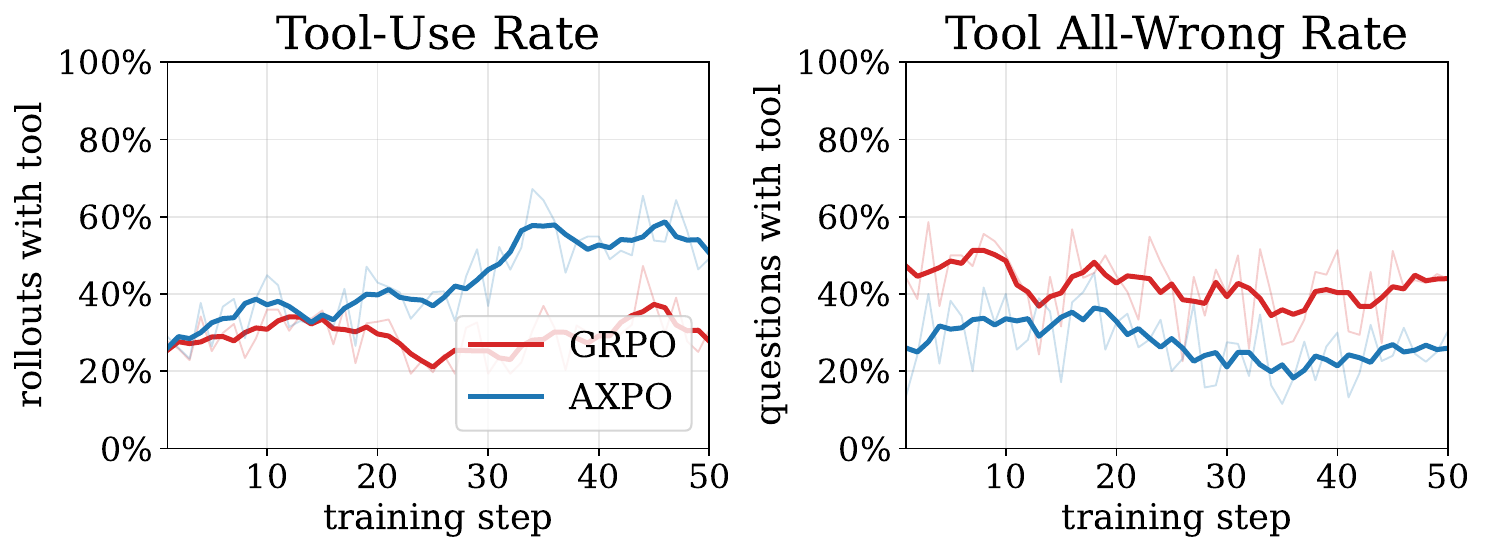}
    \end{minipage}
    \begin{minipage}[b]{0.34\linewidth}
        \centering
        \includegraphics[width=\linewidth]{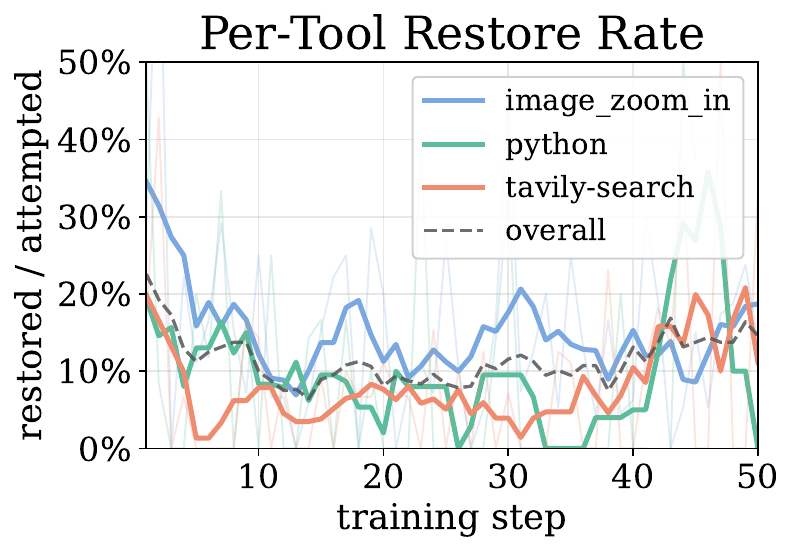}
    \end{minipage}
    \vspace{-0.1in}
    \caption{Both \gap symptoms reverse during AXPO training but stay flat under GRPO, and the recovery does happen at resampling. \textbf{Left \& Middle:} tool-use rate (top) and all-wrong rate on tool-using subgroups (bottom) over RL steps. \textbf{Right:} per-tool failed-subgroup recovery rate --- the share of all-wrong tool-using subgroups that resampling flips into a subgroup with at least one correct tool-using trajectory.}
    \vspace{-0.15in}
    \label{fig:training-dynamics}
\end{figure}

\begin{figure}[t]
    \centering
    \begin{subfigure}[b]{0.38\linewidth}
        \centering
        \includegraphics[width=\linewidth]{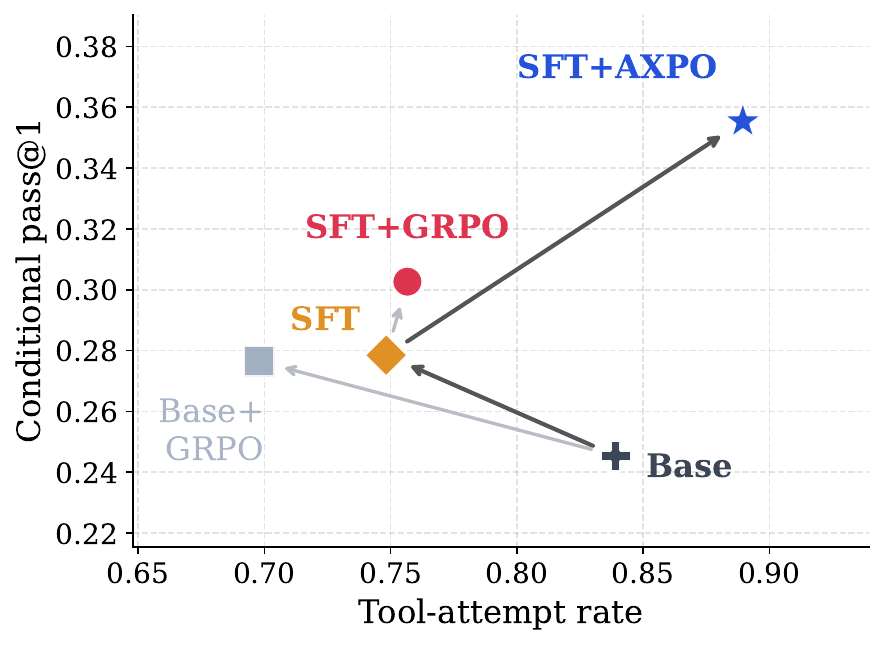}
        \vspace{-0.15in}
        \caption{Training stages on the tool-attempt rate vs. conditional pass@1 plane. Only AXPO expands both axes simultaneously.}
        \label{fig:tool-calibration}
    \end{subfigure}\hfill
    \begin{subfigure}[b]{0.60\linewidth}
        \centering
        \includegraphics[width=\linewidth]{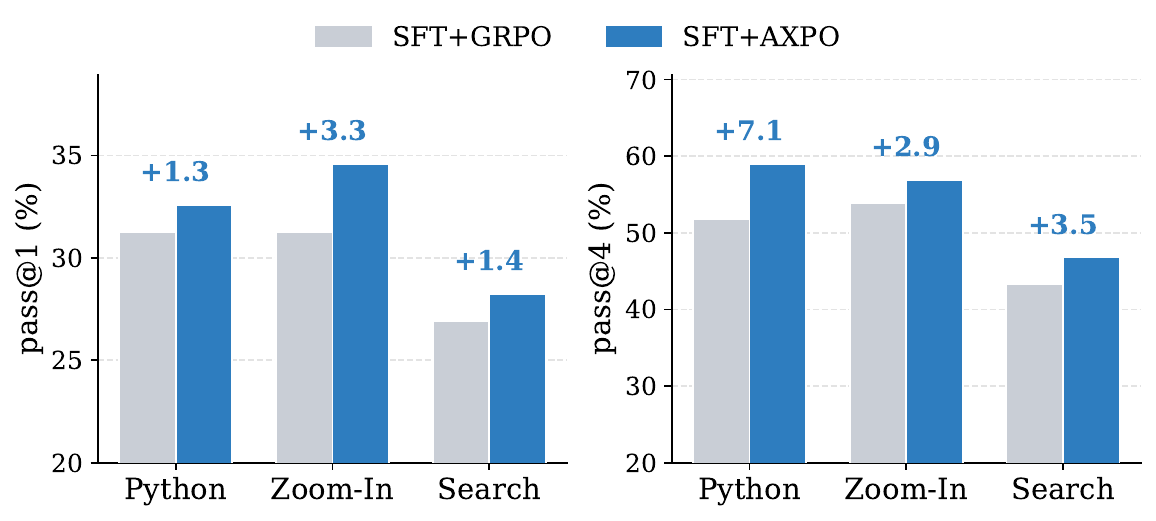}
        \vspace{-0.15in}
        \caption{Per-tool quality on the matched-tool-use subset (questions where both SFT+GRPO and SFT+AXPO invoke the tool). AXPO improves pass@1 and pass@4 across all three tool families.}
        \label{fig:per-tool-quality}
    \end{subfigure}
    \vspace{-0.05in}
    \caption{\textbf{Post-training analysis} pooled over MathVision, VisualProbe-Hard, and HR-MMSearch on 8B model. Conditional Pass@1 is measured on 715 tool-attempted questions from each benchmark.}
    \vspace{-0.2in}
    \label{fig:matched-analysis}
\end{figure}

\paragraph{Training dynamics.}
\autoref{fig:training-dynamics} (left \& middle) tracks the two diagnostic symptoms of \autoref{sec:gap} over RL steps. Under SFT\,+\,GRPO, the tool-use rate stays flat near $\sim$30\% and the all-wrong rate hovers around $\sim$40\% throughout training. Under SFT\,+\,AXPO, tool-use climbs $+28$\,pp over training and the all-wrong rate sits $\sim$17\,pp below GRPO at \emph{every} step, with tool-use rising during AXPO training while the all-wrong rate is suppressed by branching at each step, both consistent only under AXPO. The AXPO all-wrong rate is measured on the post-resampling subgroup composition (a tool-using subgroup that resampling recovers with at least one correct continuation no longer counts as all-wrong), so this persistent middle-panel gap is the per-step restore rate visualized in the right panel: \toolresample flips $\sim$12\% of all-wrong tool-using subgroups per step into subgroups with at least one correct tool-call continuation. This rate is non-trivial because resampling targets the hardest cases where every original tool-using rollout already failed, and restores learning signal so the policy is rewarded for tool use.

\paragraph{Test-time dynamics.}
More globally, \autoref{fig:tool-calibration} positions each training stage on the (tool-attempt rate, conditional Pass@1) plane measured on three benchmarks: every other stage either trades tool frequency for quality (Base\,$\to$\,GRPO or SFT) or holds tool frequency fixed while pushing only quality (SFT\,$\to$\,SFT\,+\,GRPO), while \textbf{only SFT\,+\,AXPO advances along both axes simultaneously}. The post-training matched-tool analysis (\autoref{fig:per-tool-quality}) is the per-question analysis of this joint advance: on questions where SFT\,+\,GRPO and SFT\,+\,AXPO both invoke a tool, AXPO improves Pass@1 across all three tool families, indicating the gain is in tool interaction quality.
\section{Related Works}
\label{sec:related}

\paragraph{Agentic reasoning for VLMs.}
Agentic reasoning extends the reasoning loop beyond the model's token space, allowing it to emit a tool call, receive an external output, and condition subsequent reasoning on it~\citep{ReAct, AgenticReasoning}. Tools span knowledge search~\citep{Search-o1, Search-R1}, computation~\citep{tora, TIR, retool, mathcoder, T1}, and visual perception~\citep{pixelreasoner, mini-o3, pyvision, deepeyes, agenticvision, thinkingwithimages}. Recent VLM systems combine image manipulation with code and search~\citep{mmsearch, deepeyesv2, pyvision-rl, visualsketchpad, vthinker}, and follow a two-stage SFT-then-RL recipe.

\paragraph{RL for agentic reasoning.}
On thinking-only trajectories, group-relative methods like GRPO~\citep{GRPO, DAPO, prorl} normalize rewards within $N$ rollouts and uniformly assign the resulting advantage; BroRL~\citep{brorl} analyzes how scaling $N$ closes coverage gaps. Three lines extend this family to tool-using rollouts: \emph{direct ports} that reuse outcome-reward RL without changing the sampling structure~\citep{deepeyes, deepeyesv2, rstar2, ToolRL, torl, thyme, AgenticReasoningRL, hrmmsearch}, \emph{mid-trajectory branching} that resamples the continuation after a tool call to absorb tool-output uncertainty~\citep{arpo, aepo}, and \emph{reward shaping} that adds bonus or penalty terms on tool use~\citep{pyvision-rl, otc}. We include more detailed comparison with related works in \appref{app:related-works}.

\paragraph{Our contribution.}
AXPO contrasts with each line. Against direct use of GRPO, it restores learning signal on the all-wrong tool-using subgroups identified in \autoref{sec:gap}, which uniform sampling cannot reach. Against mid-trajectory branching~\citep{arpo, aepo}, it resamples \emph{at} the tool-call boundary rather than after the tool observation, exploring the tool call itself rather than the continuation given a tool output. Against reward shaping, it acts on the rollout distribution while leaving the reward intact.

\section{Conclusion}
\label{sec:conclusion}

We identified a \emph{Thinking-Acting Gap} in group-based agentic RL, an asymmetry between thinking (the self-contained default) and tool use (a high-variance auxiliary) that manifests during training as two symptoms: tool use is attempted on a minority of rollouts and tool-using subgroups (the tool-using rollouts within a group) are disproportionately all-wrong, so the advantage at tool-call tokens is non-positive on exactly the questions where tool use would help. AXPO narrows this gap by augmenting the rollout loop with \emph{\toolresample}, which fixes the thinking prefix and concentrates exploration on the tool call itself, paired with uncertainty-based prefix selection. Across nine multimodal benchmarks and three model scales, AXPO consistently outperforms SFT\,+\,GRPO, and at 8B it surpasses the 32B Base baseline on Pass@4 with $4\times$ fewer parameters.

\paragraph{Limitations.}
Our study assumes verifiable outcome rewards for RL and trains models up to 8B parameters.
We discuss limitations and the future directions in details in \appref{app:limitations}.

\clearpage
\bibliography{reference}

\appendix
\newpage
\section{Experimental Setup}
\label{app:setup}

This section expands on \autoref{sec:setup}: training data and hyperparameters (\appref{app:training-setup}), the agentic system prompt and tool schemas (\appref{app:tools}), and per-benchmark descriptions (\appref{app:benchmarks}).

\subsection{Training}
\label{app:training-setup}

We expand on \autoref{sec:setup} with the full training data composition and hyperparameter list.

\paragraph{SFT data composition.}
SFT uses 64{,}274 trajectories drawn from three sources: ViRL~\citep{virl}, fvqa~\citep{fvqa}, and PyVision-RL~\citep{pyvision-rl}. All trajectories are generated by Qwen3-VL-32B-Thinking~\citep{Qwen3VL} acting as the teacher model. $\sim$25\% of the trajectories contain at least one tool call; the remainder solve the problem with internal reasoning alone. We retain only trajectories whose final answer is correct against the ground-truth label.

\paragraph{RL data composition.}
RL uses 15{,}591 problems obtained by filtering the SFT problem set: we remove problems that the SFT checkpoint solves on all 4 rollouts, and problems that the 32B teacher fails on all 4 rollouts. This filtering convention follows prior agentic RL work~\citep{deepeyesv2, prorl} and removes both trivial problems (zero learning signal) and unreachable problems (no positive signal achievable at this scale). We additionally include 22{,}000 hard questions from MMFineReason-hard\footnote{A question subset of the MMFineReason corpus that Qwen3-VL-4B entirely fails on 4 rollouts.} to expose the policy to problems unseen during SFT, bringing the RL training set to $\sim$37K problems.

\paragraph{Policy gradient objective.}
\label{app:clip-objective}
Both SFT\,+\,GRPO and SFT\,+\,AXPO \emph{maximize} the standard PPO-clip surrogate~\citep{PPO} regularized by a KL penalty toward a fixed reference policy $\pi_{\text{ref}}$ (the SFT-initialized checkpoint). Letting $\rho_t(\theta) = \pi_\theta(y_t \mid \vx, \vy_{<t}) / \pi_{\theta_{\text{old}}}(y_t \mid \vx, \vy_{<t})$ denote the per-token importance ratio against the rollout policy $\pi_{\theta_{\text{old}}}$ and $\hat{A}$ the trajectory-level advantage broadcast uniformly to every token of $\vy$,
\begin{gather}
\label{eq:clip-app}
    \mathcal{L}_{\text{clip}}(\vy;\, \hat{A}) \;=\; \mathbb{E}_{t}\!\big[\, \min\!\bigl(\rho_t(\theta)\, \hat{A},\; \mathrm{clip}(\rho_t(\theta),\, 1-\epsilon_{\text{low}},\, 1+\epsilon_{\text{high}})\, \hat{A}\bigr) \\
    \;-\; \beta\, D_{\mathrm{KL}}\!\left[\pi_\theta(\cdot \mid \vx, \vy_{<t}) \,\Vert\, \pi_{\text{ref}}(\cdot \mid \vx, \vy_{<t})\right] \,\big]. \nonumber
\end{gather}
The training objective is $\max_\theta\; \mathbb{E}_{\vy \sim \pi_{\theta_{\text{old}}}}\!\left[\mathcal{L}_{\text{clip}}(\vy;\, \hat{A})\right]$. GRPO instantiates $\hat{A} = A_i$ for trajectory $\tau_i$ from \autoref{sec:prelim-rl}; AXPO applies the same surrogate per stream in \autoref{eq:axpo-loss} (per-prefix advantage on continuations, recovery-indicator advantage on the source prefix). We use the asymmetric DAPO clipping range $\epsilon_{\text{low}} = 0.2,\, \epsilon_{\text{high}} = 0.4$ and KL coefficient $\beta = 1\times 10^{-3}$.

\paragraph{Hyperparameters.}
We adopt the ProRL~\citep{prorl} recipe for the RL stage. The key hyperparameters are:
\begin{itemize}[leftmargin=*, itemsep=2pt]
    \item Sampling temperature: $1.0$
    \item KL coefficient: $1\times 10^{-3}$
    \item Asymmetric clipping: $\epsilon_{\text{low}} = 0.2$, $\epsilon_{\text{high}} = 0.4$ (DAPO-style~\citep{DAPO})
    \item Group size: $N = 8$ rollouts per question
    \item Maximum number of turns per trajectory: $3$
    \item Hardware: $8 \times$ A100 80GB GPUs
\end{itemize}
All hyperparameters are held constant between SFT\,+\,GRPO and SFT\,+\,AXPO; only the rollout loop differs (see below).

For implementation of the training pipeline, we used verl~\citep{verl} and rllm~\citep{rllm} libraries.

\paragraph{AXPO-specific hyperparameters.}
AXPO adds two knobs on top of the GRPO recipe:
\begin{itemize}[leftmargin=*, itemsep=2pt]
    \item Branching budget: $r = 0.25$ of total rollout (i.e., extra continuations per training step are capped at $0.25 \cdot BN$).
    \item Per-prefix resample count: $K = 4$ continuations per selected prefix.
\end{itemize}

\subsection{System Prompt and Tool Interface}
\label{app:tools}

We use the same agentic system prompt for SFT, RL, and evaluation across all model sizes where the prompt is modified version of the system prompt of deepeyes v2~\citep{deepeyesv2}.
Tools are exposed via a unified \texttt{<tool\_call>} interface used in qwen-agent~\citep{Qwen3VL} and the model is free to invoke any subset of the three tools (\texttt{python}, \texttt{tavily-search}, \texttt{image\_zoom\_in}) on every question. 
The full text of the prompt is reproduced below.

\begin{promptbox}
You are an agent: please keep going until the user's query is completely resolved, before ending your turn and yielding back to the user. Only terminate your turn when you are sure that the problem is solved.

Solve the following problem step by step. In your reasoning process, if the answer cannot be determined, you can write Python code in a Jupyter Notebook to process the image and extract more information from it. The stdout and stderr content, along with the images generated by ``\texttt{plt.show()}'' will be returned to better assist with the user query.

You MUST use the python tool to analyze or transform images whenever it could improve your understanding. This includes but is not limited to zooming in, rotating, adjusting contrast, computing statistics, or isolating features. If you find you lack sufficient knowledge to confidently answer the question, you MUST conduct search to thoroughly seek the internet for information. No matter how complex the query, you will not give up until you find the corresponding information. You can use the tavily-search tool, which will search the web and return relevant information based on your query.

\textbf{Notes for tools:}

\textbf{\#\# image\_zoom\_in tool}
\begin{enumerate}[leftmargin=*, itemsep=0pt, topsep=2pt]
    \item \textbf{image\_zoom\_in} crops and magnifies a specific region of the image so you can inspect fine details.
    \item When you need to look at a specific region more closely, prefer \textbf{image\_zoom\_in} over writing python code to crop --- it is simpler, faster, and returns a high-resolution view directly.
    \item After a zoom-in, the cropped image is appended to \texttt{\_pil\_images} in the python tool (the observation will tell you the exact index), so you can further process it with code if needed.
\end{enumerate}

\textbf{\#\# python tool}
\begin{enumerate}[leftmargin=*, itemsep=0pt, topsep=2pt]
    \item \textbf{python} runs in a stateful Jupyter notebook kernel --- variables, imports, and state persist across calls.
    \item The conversation images are pre-loaded as \texttt{\_pil\_images} (list of PIL Image objects) and \texttt{\_images} (list of file paths). Images produced by other tools (e.g.\ \texttt{image\_zoom\_in}) are also appended to these lists.
    \item Use \texttt{plt.show()} to visualize processed images; generated plots are automatically captured and returned.
    \item Available packages include numpy, scipy, sympy, matplotlib, PIL/Pillow, and the standard library.
    \item \textbf{Always use Python for non-trivial computation} --- trigonometric calculations, multi-step arithmetic, equation solving, and area/volume formulas. Do NOT compute these mentally; write code to ensure numerical accuracy.
    \item Your program should always return in finite time. Do not write infinite loops.
    \item Writing file to disk is not allowed.
\end{enumerate}

\textbf{\#\# tavily-search tool}
\begin{enumerate}[leftmargin=*, itemsep=0pt, topsep=2pt]
    \item You MUST engage in many interactions, delving deeply into the topic to explore all possible aspects until a satisfactory answer is found.
    \item Before presenting a Final Answer, you will \textbf{cross-check} and \textbf{validate the information} you've gathered to confirm its accuracy and reliability.
    \item You will carefully analyze each information source to ensure that all data is current, relevant, and from credible origins.
\end{enumerate}

\textbf{\#\# general}

You MUST think and plan before each function call, and reflect on the outcomes of the previous function calls. DO NOT do this entire process by making function calls only, as this can impair your ability to solve the problem and think insightfully.

Remember to place the final answer in the last part using the format:
\begin{center}
\texttt{<answer>}\\
\texttt{\textbackslash boxed\{"The final answer goes here."\}}\\
\texttt{</answer>}
\end{center}
\end{promptbox}

\paragraph{Tool schemas.}
Each tool is registered with the model using the OpenAI function-calling schema (\texttt{name}, \texttt{description}, JSON \texttt{parameters}) and is concatenated with the system prompt above at every step. The three schemas in full are below.

\paragraph{\texttt{python}.}
\begin{Verbatim}[fontsize=\small, breaklines=true, breakanywhere=true, breakindent=2em]
{
  "type": "function",
  "function": {
    "name": "python",
    "description": "Execute Python code in a persistent Jupyter kernel. State (variables, imports) is preserved across calls. Returns stdout/stderr and captured images. Conversation images are pre-loaded as `_pil_images` (list of PIL.Image) and `_images` (list of file paths). Use `_pil_images[0]` to access the first image directly - do NOT use plt.imread() or open files manually.",
    "parameters": {
      "type": "object",
      "properties": {
        "code": {
          "type": "string",
          "description": "Python code to execute in the sandbox environment."
        },
        "timeout": {
          "type": "integer",
          "description": "Maximum execution time in seconds before timing out",
          "default": 12
        }
      },
      "required": ["code"]
    }
  }
}
\end{Verbatim}

\paragraph{\texttt{image\_zoom\_in}.}
\begin{Verbatim}[fontsize=\small, breaklines=true, breakanywhere=true, breakindent=2em]
{
  "type": "function",
  "function": {
    "name": "image_zoom_in",
    "description": "Zoom in on a specific region of an image by cropping it based on a bounding box (bbox) and an optional object label",
    "parameters": {
      "type": "object",
      "properties": {
        "bbox_2d": {
          "type": "array",
          "items": {"type": "number"},
          "minItems": 4,
          "maxItems": 4,
          "description": "The bounding box of the region to zoom in, as [x1, y1, x2, y2] in 0-1000 relative coordinates, where (x1, y1) is the top-left corner and (x2, y2) is the bottom-right corner"
        },
        "label": {
          "type": "string",
          "description": "The name or label of the object in the specified bounding box"
        },
        "img_idx": {
          "type": "number",
          "description": "The index of the image to zoom in on (starting from 0)",
          "default": 0
        }
      },
      "required": ["bbox_2d", "label"]
    }
  }
}
\end{Verbatim}

\paragraph{\texttt{tavily-search}.}
\begin{Verbatim}[fontsize=\small, breaklines=true, breakanywhere=true, breakindent=2em]
{
  "type": "function",
  "function": {
    "name": "tavily-search",
    "description": "Search the web for information on a specific query",
    "parameters": {
      "type": "object",
      "properties": {
        "query": {
          "type": "string",
          "description": "The search query"
        },
        "max_results": {
          "type": "integer",
          "description": "Maximum number of search results to return. Defaults to 5."
        }
      },
      "required": ["query"]
    }
  }
}
\end{Verbatim}

\paragraph{Search domain blacklist.}
To prevent answer leakage during training and evaluation, \texttt{tavily-search} is configured with the API-level parameter \texttt{exclude\_domains = ["huggingface.co"]}, which filters out the site most likely to index benchmark questions or solutions. The exclusion is enforced uniformly across training and evaluation rollouts.

\subsection{Benchmark Details}
\label{app:benchmarks}

We evaluate on nine multimodal benchmarks grouped by which tool family is most expected to help: \textbf{Reasoning} (Python interpreter), \textbf{Perception} (image zoom-in), and \textbf{Search} (web search). All benchmarks are evaluated in the agentic setting with the same tool interface used in training; the model is free to use any subset of the three tools or none at all on every question.

\paragraph{Reasoning.} These benchmarks evaluate multi-step reasoning over text, math, and figures, where a Python interpreter is the natural tool for symbolic and numerical computation.
\begin{itemize}[leftmargin=*, itemsep=2pt]
    \item \textbf{MathVision}~\citep{mathvision}: competition-style math problems paired with diagrams (geometry, algebra, combinatorics). Answers are short numerical or symbolic expressions; we use the public mini test set containing 304 questions.
    \item \textbf{DynaMath}~\citep{dynamath}: a benchmark of programmatically generated math questions with multiple parametric variants per seed problem. We evaluate on a single-variant slice (501 seed problems). The benchmark  complements MathVision as it provides additional algorithmically generated problems spanning plane geometry, statistics, and other domains with diverse difficulty.
    \item \textbf{Math-VR}~\citep{mathvr}: a mathematical visual reasoning benchmark introduced with CodePlot-CoT, in which problems are paired with diagrams (figures, plots, geometric constructions) that the model must inspect to compute the answer. The benchmark targets the regime where the visual content is load-bearing for the math derivation rather than decorative.
\end{itemize}

\paragraph{Perception.} These benchmarks contain high-resolution images with information that is hard to read at the default vision encoder resolution. The image zoom-in tool is the natural recovery mechanism.
\begin{itemize}[leftmargin=*, itemsep=2pt]
    \item \textbf{V$^{\star}$}~\citep{vstar}: a fine-grained visual-search benchmark in which the answer depends on a small object embedded in a complex scene. Designed to expose the failure of monolithic image encoders on high-resolution detail.
    \item \textbf{VisualProbe}~\citep{mini-o3}: a multimodal probing benchmark introduced with mini-o3, where each question targets a specific small region (text, sign, label) within a high-resolution scene. We use the \emph{hard} subset, which selects questions Qwen3-VL-Thinking-8B fails without tools.
    \item \textbf{HR-Bench-4K and HR-Bench-8K}~\citep{HRBench}: high-resolution single-image perception at $4096^2$ and $8192^2$ resolutions, respectively. Questions probe small text, distant objects, and fine textures; the two splits share a common question pool with images rescaled to the target resolution.
\end{itemize}

\paragraph{Search.} These benchmarks require external knowledge that is unlikely to be in the model's parametric memory or that requires multi-hop lookup. The Tavily web search API is the natural tool.
\begin{itemize}[leftmargin=*, itemsep=2pt]
    \item \textbf{HR-MMSearch}~\citep{hrmmsearch}: high-resolution multimodal search questions whose answers are composed by chaining two or more web lookups (e.g., ``in which year was the mayor of the city where this fashion designer was born''). The benchmark stresses both perception (identifying entities from the image) and multi-hop search composition.
    \item \textbf{MMSearch}~\citep{mmsearch}: an earlier multimodal search benchmark covering current-events questions whose answers depend on web content released after the model's training cutoff. Predominantly single-hop, but answers are often time-sensitive.
\end{itemize}

\paragraph{Evaluation protocol.} For all benchmarks, we draw four rollouts per question and report Pass@1 (per-question accuracy averaged over four rollouts) and Pass@4 (a question is correct if any of the four rollouts is correct). Final-answer matching against the gold label is performed by an exact match for math and multiple-choice question and F1 score for search questions. For benchmarks that are hard to evaluate with exact match (e.g., MMSearch), we use a llm-as-a-judge using gpt-5-mini model. 

\paragraph{Evaluation hyperparameters.}
\begin{itemize}[leftmargin=*, itemsep=2pt]
    \item Sampling: temperature $0.7$, top-$p$ $0.8$, top-$k$ $20$, presence penalty $1.5$
    \item Rollouts per question: $N = 4$ (used for Pass@1 / Pass@4)
    \item Maximum agentic turns per trajectory: $5$
    \item Maximum response length: $32{,}768$ tokens
\end{itemize}

\section{Additional Experimental Results}
\label{app:additional-results}

This section collects supplementary results that complement the main Experiments: tool utilization rates at evaluation (\appref{app:tool-util}), per-benchmark Pass@4 numbers (\appref{app:pass4-table}), per-rollout variance for Pass@1 (\appref{app:pass1-std}), a stage-by-stage decomposition of the SFT-then-RL recipe (\appref{app:decomp-recipe}), generalization to an unseen tool (\appref{app:image-search-gen}), per-row analysis of the ablation table (\appref{app:ablation-details}), per-row analysis of the comparison table (\appref{app:comparison-details}), a comparison against prior agentic VLM systems (\appref{app:prior-agentic-vlms}), and a post-hoc justification of confidence as the prefix-ranking signal (\appref{app:conf-vs-entropy}).

\subsection{Tool Utilization Rates at Evaluation}
\label{app:tool-util}
\begin{table*}[t]
    \caption{\textbf{Tool utilization rate (\%, $\uparrow$)} per benchmark, mirroring \autoref{tab:main-p1}. Tool utilization is the fraction of evaluation rollouts that emit at least one \toolcall on each benchmark. Non-agent rollouts (no tools) have tool utilization $0$ by construction and are omitted; ``Base'' is the inference-only baseline with tools enabled. Best per column within each size block in \textbf{bold}.}
    \label{tab:tool-util}
    \centering
    \scriptsize
    \resizebox{\textwidth}{!}{
    \begin{tabular}{l ccc cccc cc >{\columncolor{mygray}}c}
        \toprule
        \multicolumn{1}{c}{} &
        \multicolumn{3}{c}{\textbf{Reasoning}} &
        \multicolumn{4}{c}{\textbf{Perception}} &
        \multicolumn{2}{c}{\textbf{Search}} &
        \multicolumn{1}{c}{\multirow{2}{*}[-3pt]{\textbf{Avg.}}} \\
        \cmidrule(lr){2-4} \cmidrule(lr){5-8} \cmidrule(lr){9-10}
        \textbf{Method} &
        \shortstack{Math\\Vision} & \shortstack{Dyna\\Math} & \shortstack{Math\\-VR} &
        V$^{\star}$ & \shortstack{Visual\\Probe} & \shortstack{HRBen.\\-4K} & \shortstack{HRBen.\\-8K} &
        \shortstack{HR-MM\\Search} & \shortstack{MM\\Search} & \multicolumn{1}{c}{} \\
        \midrule

        \multicolumn{11}{l}{\textit{Qwen3-VL-2B-Thinking}} \\
        \midrule
        Base       & 78.3 & 86.4 & 76.7 & 94.8 & 99.1 & 87.5 & 93.0 & 92.8 & 97.3 & 89.5 \\
        GRPO       & \textbf{96.1} & \textbf{99.0} & \textbf{93.2} & 96.9 & 100.0 & 90.0 & \textbf{94.0} & \textbf{100.0} & \textbf{100.0} & \textbf{96.6} \\
        SFT        & 38.8 & 34.3 & 8.0  & 95.8 & 100.0 & 79.5 & 80.5 & 91.5 & 95.0 & 69.3 \\
        \quad+\,GRPO  & 54.6 & 30.7 & 17.1 & 90.1 & 100.0 & 72.0 & 80.5 & 94.1 & 97.0 & 70.7 \\
        \rowcolor{myblue}
        \quad+\,\textbf{AXPO}\,\scalebox{0.8}{\textit{(Ours)}}
                  & 63.8 & 63.9 & 27.1 & \textbf{99.0} & 100.0 & \textbf{93.0} & 93.0 & 98.7 & 99.0 & 81.9 \\
        \midrule

        \multicolumn{11}{l}{\textit{Qwen3-VL-4B-Thinking}} \\
        \midrule
        Base       & 25.7 & 17.0 & 1.6  & 94.8 & 99.1 & 79.5 & 85.5 & 85.2 & 95.7 & 64.9 \\
        GRPO       & 2.3  & 2.4  & 0.0  & 65.4 & 90.6 & 36.5 & 42.5 & 35.1 & 63.0 & 37.5 \\
        SFT        & 53.9 & 38.9 & 10.6 & 94.8 & 100.0 & 75.5 & 83.5 & 93.1 & 97.7 & 72.0 \\
        \quad+\,GRPO  & 38.8 & 27.7 & 8.6  & 80.6 & 100.0 & 64.0 & 74.5 & 91.5 & 95.7 & 64.6 \\
        \rowcolor{myblue}
        \quad+\,\textbf{AXPO}\,\scalebox{0.8}{\textit{(Ours)}}
                  & \textbf{73.0} & \textbf{96.2} & \textbf{13.3} & \textbf{100.0} & 100.0 & \textbf{100.0} & \textbf{100.0} & \textbf{100.0} & \textbf{100.0} & \textbf{86.9} \\
        \midrule

        \multicolumn{11}{l}{\textit{Qwen3-VL-8B-Thinking}} \\
        \midrule
        Base       & 62.2 & 50.7 & 22.5 & 94.8 & 100.0 & 84.5 & 85.0 & 94.4 & \textbf{98.7} & 77.0 \\
        GRPO       & 38.2 & 22.0 & 7.8  & 87.4 & 99.1 & 72.0 & 75.5 & 91.1 & 97.3 & 65.6 \\
        SFT        & 47.7 & 34.1 & 9.8  & 90.6 & 99.1 & 74.0 & 76.0 & 93.4 & 96.7 & 69.0 \\
        \quad+\,GRPO  & 46.1 & 30.7 & 7.6  & 92.1 & 99.1 & 73.0 & 78.0 & 97.0 & 97.7 & 69.0 \\
        \rowcolor{myblue}
        \quad+\,\textbf{AXPO}\,\scalebox{0.8}{\textit{(Ours)}}
                  & \textbf{75.7} & \textbf{63.9} & \textbf{29.1} & \textbf{99.0} & 100.0 & \textbf{90.0} & \textbf{92.0} & \textbf{98.4} & 98.3 & \textbf{82.9} \\
        \midrule

        \multicolumn{11}{l}{\textit{Qwen3-VL-32B-Thinking} (inference-only baseline)} \\
        \midrule
        Base       & 51.6 & 36.5 & 15.1 & 94.8 & 100.0 & 75.0 & 77.5 & 94.8 & 97.3 & 71.4 \\

        \bottomrule
    \end{tabular}
    }
\end{table*}

\autoref{tab:tool-util} reports tool utilization (the fraction of evaluation rollouts that emit at least one \toolcall) on each of the nine benchmarks, mirroring the structure of \autoref{tab:main-p1}. Across all three trainable scales (2B/4B/8B), SFT\,+\,AXPO produces the highest average tool utilization among trained methods (\toolUtilAxpoAvg), and the gap to SFT\,+\,GRPO grows with size only on benchmarks where tool use is load-bearing (e.g., \toolUtilHardReasoning{} at 8B: \toolUtilHardReasoningGap\,pp), while on tool-saturated benchmarks (V$^{\star}$, VisualProbe, MMSearch) both methods sit near $100\%$. Thus AXPO improves both tool-use frequency \emph{and}, as \autoref{fig:per-tool-quality} suggests, conditional tool-use quality. The matched-tool analysis does not rule out frequency effects but rather shows that gains remain even on the subset of questions where both SFT\,+\,GRPO and SFT\,+\,AXPO invoke the same tool family.

\subsection{Per-Benchmark Pass@4 Results}
\label{app:pass4-table}
\begin{table*}[t]
    \caption{\small \textbf{Main results: Pass@4} (\%, $\uparrow$) on \textbf{nine multimodal benchmarks}, grouped into three categories: Reasoning, Perception, and Search. All methods use \textbf{Qwen3-VL-Thinking}~\citep{Qwen3VL} at four sizes (2B, 4B, 8B, 32B). Base (tools, no training) is the inference-only baseline; the Non-agent (no tools) baseline is reported separately in \autoref{tab:non-agent-vs-base}. Best result per column within each size block is \textbf{bolded}; second-best is \underline{underlined}. \textit{$\Delta$} rows show AXPO's gain over SFT\,+\,GRPO, the strongest comparable training recipe where AXPO outperforms it in most of evaluations. Pass@1 results are reported in \autoref{tab:main-p1}.}
    \vspace{-0.1in}
    \label{tab:main-p4}
    \centering
    \small
    \resizebox{\textwidth}{!}{
    \begin{tabular}{l ccc cccc cc >{\columncolor{mygray}}c}
        \toprule
        \multicolumn{1}{c}{} &
        \multicolumn{3}{c}{\textbf{Reasoning}} &
        \multicolumn{4}{c}{\textbf{Perception}} &
        \multicolumn{2}{c}{\textbf{Search}} &
        \multicolumn{1}{c}{\multirow{2}{*}[-3pt]{\textbf{Avg.}}} \\
        \cmidrule(lr){2-4} \cmidrule(lr){5-8} \cmidrule(lr){9-10}
        \textbf{Method} &
        \shortstack{Math\\Vision} & \shortstack{Dyna\\Math} & \shortstack{Math\\-VR} &
        V$^{\star}$ & \shortstack{Visual\\Probe} & \shortstack{HRBen.\\-4K} & \shortstack{HRBen.\\-8K} &
        \shortstack{HR-MM\\Search} & \shortstack{MM\\Search} & \multicolumn{1}{c}{} \\
        \midrule

        \multicolumn{11}{l}{\textit{Qwen3-VL-2B-Thinking}} \\
        \midrule
        Base
            & 32.6 & 76.6 & 28.9 & 81.7 & 15.1 & 79.0 & 77.0 & 18.7 & 46.3 & 50.7 \\
        GRPO
            & 53.3 & 78.2 & 52.0 & 80.6 & 43.4 & 75.5 & 72.5 & \underline{29.5} & 54.7 & 60.0 \\
        SFT
            & 53.0 & \underline{80.2} & 49.2 & \underline{91.1} & \underline{50.0} & \underline{86.5} & \textbf{88.0} & 28.5 & 53.7 & 64.5 \\
        \quad+\,GRPO
            & \underline{55.9} & \underline{80.2} & \underline{58.0} & \textbf{94.2} & \textbf{60.4} & 83.5 & \underline{85.5} & 28.2 & \underline{55.7} & \underline{66.8} \\
        \rowcolor{myblue}
        \quad+\,\textbf{AXPO}\,\scalebox{0.8}{\textit{(Ours)}}
            & \textbf{61.2} & \textbf{83.6} & \textbf{60.0} & \textbf{94.2} & \textbf{60.4} & \textbf{88.5} & \textbf{88.0} & \textbf{33.4} & \textbf{57.3} & \textbf{69.6} \\
        \rowcolor{myblue!50}
        \textit{\;\;$\Delta$ vs.\ SFT\,+\,GRPO}
            & \dpos{5.3} & \dpos{3.4} & \dpos{2.0} & \dpos{0.0} & \dpos{0.0} & \dpos{5.0} & \dpos{2.5} & \dpos{5.2} & \dpos{1.6} & \dposavg{2.8} \\
        \midrule

        \multicolumn{11}{l}{\textit{Qwen3-VL-4B-Thinking}} \\
        \midrule
        Base
            & 64.8 & 82.6 & 65.1 & \underline{95.8} & 44.3 & 83.5 & 81.0 & 27.5 & 55.7 & 66.7 \\
        GRPO
            & 69.1 & 84.2 & 70.5 & 92.7 & 48.1 & 87.0 & 85.5 & 22.0 & 42.0 & 66.8 \\
        SFT
            & 69.4 & \underline{86.4} & 69.5 & 95.3 & 51.9 & \underline{89.5} & \underline{88.0} & \textbf{35.7} & \textbf{59.7} & \underline{71.7} \\
        \quad+\,GRPO
            & \underline{69.7} & 85.8 & \underline{71.9} & \textbf{97.9} & \underline{57.5} & 88.0 & 84.5 & \underline{33.1} & \underline{57.0} & \underline{71.7} \\
        \rowcolor{myblue}
        \quad+\,\textbf{AXPO}\,\scalebox{0.8}{\textit{(Ours)}}
            & \textbf{74.3} & \textbf{87.0} & \textbf{72.1} & \textbf{97.9} & \textbf{65.1} & \textbf{91.0} & \textbf{90.0} & 32.1 & \underline{57.0} & \textbf{74.1} \\
        \rowcolor{myblue!50}
        \textit{\;\;$\Delta$ vs.\ SFT\,+\,GRPO}
            & \dpos{4.6} & \dpos{1.2} & \dpos{0.2} & \dpos{0.0} & \dpos{7.6} & \dpos{3.0} & \dpos{5.5} & \dneg{1.0} & \dpos{0.0} & \dposavg{2.3} \\
        \midrule

        \multicolumn{11}{l}{\textit{Qwen3-VL-8B-Thinking}} \\
        \midrule
        Base
            & 63.5 & 85.2 & 68.3 & 92.7 & 55.7 & 85.5 & 82.5 & 33.8 & 59.0 & 69.6 \\
        GRPO
            & 69.7 & 84.6 & 70.3 & 94.2 & 62.3 & 87.0 & 82.0 & 34.4 & 60.0 & 71.6 \\
        SFT
            & 71.4 & \underline{86.0} & 72.1 & 93.7 & 57.5 & \textbf{90.5} & \underline{87.0} & 36.4 & 55.3 & 72.2 \\
        \quad+\,GRPO
            & \underline{75.0} & \textbf{87.0} & \textbf{74.3} & \textbf{96.9} & \underline{63.2} & 87.5 & 84.5 & \underline{38.0} & \underline{59.3} & \underline{74.0} \\
        \rowcolor{myblue}
        \quad+\,\textbf{AXPO}\,\scalebox{0.8}{\textit{(Ours)}}
            & \textbf{75.7} & \textbf{87.0} & \underline{74.1} & \underline{95.3} & \textbf{67.9} & \underline{90.0} & \textbf{89.0} & \textbf{42.0} & \textbf{61.0} & \textbf{75.8} \\
        \rowcolor{myblue!50}
        \textit{\;\;$\Delta$ vs.\ SFT\,+\,GRPO}
            & \dpos{0.7} & \dpos{0.0} & \dneg{0.2} & \dneg{1.6} & \dpos{4.7} & \dpos{2.5} & \dpos{4.5} & \dpos{4.0} & \dpos{1.7} & \dposavg{1.8} \\
        \midrule

        \multicolumn{11}{l}{\textit{Qwen3-VL-32B-Thinking} (inference-only baseline)} \\
        \midrule
        Base
            & 73.0 & 89.4 & 73.9 & 97.4 & 64.2 & 90.5 & 90.0 & 38.0 & 59.3 & 75.1 \\

        \bottomrule
    \end{tabular}
    }
\vspace{-0.2in}
\end{table*}

The main manuscript reports Pass@1 in \autoref{tab:main-p1}; \autoref{tab:main-p4} reports the corresponding Pass@4 numbers across the same nine benchmarks and four model sizes, with identical method ordering and bold/underline conventions. The two tables share an Average column, which is the basis for the headline numbers in \autoref{sec:main-results} and the per-stage decomposition below.

\subsection{Pass@1 Variance Across Independent Rollouts}
\label{app:pass1-std}
\begin{table*}[t]
    \caption{\textbf{Pass@1 mean\,$\pm$\,std} (\%, $\uparrow$) across \textbf{four independent rollouts} on nine multimodal benchmarks. Subscripts denote standard deviation across the four rollouts. Variance is comparable across methods, with most std values in the $0.4$--$3.6$\,pp range; the column-averaged std for SFT\,+\,AXPO is $1.4$\,pp at 4B and $1.2$\,pp at 8B.}
    \vspace{-0.1in}
    \label{tab:main-p1-std}
    \centering
    \scriptsize
    \resizebox{\textwidth}{!}{
    \begin{tabular}{l ccc cccc cc >{\columncolor{mygray}}c}
        \toprule
        \multicolumn{1}{c}{} &
        \multicolumn{3}{c}{\textbf{Reasoning}} &
        \multicolumn{4}{c}{\textbf{Perception}} &
        \multicolumn{2}{c}{\textbf{Search}} &
        \multicolumn{1}{c}{\multirow{2}{*}[-3pt]{\textbf{Avg.}}} \\
        \cmidrule(lr){2-4} \cmidrule(lr){5-8} \cmidrule(lr){9-10}
        \textbf{Method} &
        \shortstack{Math\\Vision} & \shortstack{Dyna\\Math} & \shortstack{Math\\-VR} &
        V$^{\star}$ & \shortstack{Visual\\Probe} & \shortstack{HRBen.\\-4K} & \shortstack{HRBen.\\-8K} &
        \shortstack{HR-MM\\Search} & \shortstack{MM\\Search} & \multicolumn{1}{c}{} \\
        \midrule

        \multicolumn{11}{l}{\textit{Qwen3-VL-2B-Thinking}} \\
        \midrule
        Base
            & 18.6\,\textsubscript{$\pm$1.8} & 60.5\,\textsubscript{$\pm$1.2} & 16.9\,\textsubscript{$\pm$1.5}
            & 63.0\,\textsubscript{$\pm$2.9} & 5.0\,\textsubscript{$\pm$1.4} & 64.6\,\textsubscript{$\pm$0.9} & 55.9\,\textsubscript{$\pm$0.5}
            & 8.2\,\textsubscript{$\pm$1.5} & 27.7\,\textsubscript{$\pm$1.4}
            & 35.6\,\textsubscript{$\pm$1.5} \\
        GRPO
            & 33.7\,\textsubscript{$\pm$1.0} & 64.9\,\textsubscript{$\pm$0.8} & 34.6\,\textsubscript{$\pm$1.0}
            & 67.1\,\textsubscript{$\pm$1.8} & 23.8\,\textsubscript{$\pm$3.4} & 65.0\,\textsubscript{$\pm$1.3} & 59.4\,\textsubscript{$\pm$1.8}
            & 18.4\,\textsubscript{$\pm$1.3} & 41.4\,\textsubscript{$\pm$0.5}
            & 45.4\,\textsubscript{$\pm$1.4} \\
        SFT
            & 28.9\,\textsubscript{$\pm$1.3} & 64.1\,\textsubscript{$\pm$1.6} & 30.2\,\textsubscript{$\pm$1.0}
            & 75.9\,\textsubscript{$\pm$2.4} & 24.5\,\textsubscript{$\pm$2.8} & 73.1\,\textsubscript{$\pm$1.1} & 68.8\,\textsubscript{$\pm$1.6}
            & 14.5\,\textsubscript{$\pm$1.7} & 35.9\,\textsubscript{$\pm$1.4}
            & 46.2\,\textsubscript{$\pm$1.7} \\
        \quad+\,GRPO
            & 34.8\,\textsubscript{$\pm$1.3} & 66.4\,\textsubscript{$\pm$1.0} & 35.8\,\textsubscript{$\pm$2.2}
            & 81.7\,\textsubscript{$\pm$3.2} & 36.1\,\textsubscript{$\pm$1.5} & 74.2\,\textsubscript{$\pm$0.8} & 70.6\,\textsubscript{$\pm$2.1}
            & 15.7\,\textsubscript{$\pm$1.0} & 40.6\,\textsubscript{$\pm$1.2}
            & 50.7\,\textsubscript{$\pm$1.6} \\
        \rowcolor{myblue}
        \quad+\,\textbf{AXPO}\,\scalebox{0.8}{\textit{(Ours)}}
            & 35.0\,\textsubscript{$\pm$1.1} & 66.9\,\textsubscript{$\pm$2.2} & 38.9\,\textsubscript{$\pm$1.0}
            & 81.3\,\textsubscript{$\pm$3.1} & 35.1\,\textsubscript{$\pm$5.0} & 77.9\,\textsubscript{$\pm$1.9} & 72.4\,\textsubscript{$\pm$1.6}
            & 18.1\,\textsubscript{$\pm$0.4} & 40.2\,\textsubscript{$\pm$1.7}
            & 51.8\,\textsubscript{$\pm$2.0} \\
        \midrule

        \multicolumn{11}{l}{\textit{Qwen3-VL-4B-Thinking}} \\
        \midrule
        Base
            & 48.0\,\textsubscript{$\pm$1.2} & 76.6\,\textsubscript{$\pm$0.7} & 50.8\,\textsubscript{$\pm$0.5}
            & 80.6\,\textsubscript{$\pm$1.2} & 24.8\,\textsubscript{$\pm$2.3} & 73.6\,\textsubscript{$\pm$1.2} & 67.5\,\textsubscript{$\pm$2.0}
            & 16.9\,\textsubscript{$\pm$0.4} & 37.8\,\textsubscript{$\pm$1.5}
            & 53.0\,\textsubscript{$\pm$1.2} \\
        GRPO
            & 52.5\,\textsubscript{$\pm$2.3} & 76.2\,\textsubscript{$\pm$1.0} & 57.9\,\textsubscript{$\pm$1.1}
            & 82.7\,\textsubscript{$\pm$2.8} & 29.7\,\textsubscript{$\pm$1.9} & 75.5\,\textsubscript{$\pm$1.3} & 71.1\,\textsubscript{$\pm$1.5}
            & 12.7\,\textsubscript{$\pm$1.2} & 23.8\,\textsubscript{$\pm$1.7}
            & 53.6\,\textsubscript{$\pm$1.6} \\
        SFT
            & 49.1\,\textsubscript{$\pm$1.3} & 76.3\,\textsubscript{$\pm$1.0} & 55.3\,\textsubscript{$\pm$1.2}
            & 87.0\,\textsubscript{$\pm$1.9} & 34.7\,\textsubscript{$\pm$1.7} & 79.5\,\textsubscript{$\pm$1.3} & 74.1\,\textsubscript{$\pm$1.0}
            & 20.6\,\textsubscript{$\pm$1.7} & 41.5\,\textsubscript{$\pm$1.5}
            & 57.6\,\textsubscript{$\pm$1.4} \\
        \quad+\,GRPO
            & 51.6\,\textsubscript{$\pm$0.6} & 77.2\,\textsubscript{$\pm$0.4} & 57.4\,\textsubscript{$\pm$1.2}
            & 90.2\,\textsubscript{$\pm$1.1} & 38.0\,\textsubscript{$\pm$2.1} & 78.8\,\textsubscript{$\pm$0.8} & 73.8\,\textsubscript{$\pm$1.5}
            & 20.0\,\textsubscript{$\pm$1.8} & 42.3\,\textsubscript{$\pm$0.8}
            & 58.8\,\textsubscript{$\pm$1.2} \\
        \rowcolor{myblue}
        \quad+\,\textbf{AXPO}\,\scalebox{0.8}{\textit{(Ours)}}
            & 52.8\,\textsubscript{$\pm$1.4} & 76.4\,\textsubscript{$\pm$0.5} & 57.2\,\textsubscript{$\pm$0.9}
            & 88.9\,\textsubscript{$\pm$1.1} & 43.6\,\textsubscript{$\pm$2.6} & 80.9\,\textsubscript{$\pm$1.7} & 78.2\,\textsubscript{$\pm$1.4}
            & 20.1\,\textsubscript{$\pm$1.3} & 43.3\,\textsubscript{$\pm$1.4}
            & 60.2\,\textsubscript{$\pm$1.4} \\
        \midrule

        \multicolumn{11}{l}{\textit{Qwen3-VL-8B-Thinking}} \\
        \midrule
        Base
            & 47.1\,\textsubscript{$\pm$2.0} & 75.9\,\textsubscript{$\pm$0.9} & 54.9\,\textsubscript{$\pm$1.1}
            & 77.7\,\textsubscript{$\pm$2.2} & 31.8\,\textsubscript{$\pm$1.8} & 72.8\,\textsubscript{$\pm$1.7} & 66.1\,\textsubscript{$\pm$2.2}
            & 21.0\,\textsubscript{$\pm$1.0} & 42.7\,\textsubscript{$\pm$1.7}
            & 54.4\,\textsubscript{$\pm$0.6} \\
        GRPO
            & 54.4\,\textsubscript{$\pm$0.7} & 77.2\,\textsubscript{$\pm$0.4} & 58.9\,\textsubscript{$\pm$1.4}
            & 85.7\,\textsubscript{$\pm$1.3} & 40.1\,\textsubscript{$\pm$3.2} & 77.1\,\textsubscript{$\pm$2.2} & 73.9\,\textsubscript{$\pm$0.5}
            & 20.6\,\textsubscript{$\pm$1.7} & 45.1\,\textsubscript{$\pm$0.6}
            & 59.2\,\textsubscript{$\pm$1.3} \\
        SFT
            & 50.9\,\textsubscript{$\pm$0.7} & 76.8\,\textsubscript{$\pm$0.6} & 58.4\,\textsubscript{$\pm$1.9}
            & 84.8\,\textsubscript{$\pm$2.0} & 38.4\,\textsubscript{$\pm$3.1} & 79.5\,\textsubscript{$\pm$0.6} & 74.4\,\textsubscript{$\pm$2.3}
            & 23.0\,\textsubscript{$\pm$1.6} & 40.6\,\textsubscript{$\pm$2.2}
            & 58.5\,\textsubscript{$\pm$1.7} \\
        \quad+\,GRPO
            & 55.3\,\textsubscript{$\pm$1.8} & 78.2\,\textsubscript{$\pm$1.0} & 60.4\,\textsubscript{$\pm$1.5}
            & 87.7\,\textsubscript{$\pm$2.4} & 40.1\,\textsubscript{$\pm$3.6} & 79.5\,\textsubscript{$\pm$1.1} & 74.9\,\textsubscript{$\pm$1.1}
            & 24.4\,\textsubscript{$\pm$1.2} & 44.0\,\textsubscript{$\pm$0.4}
            & 60.5\,\textsubscript{$\pm$0.5} \\
        \rowcolor{myblue}
        \quad+\,\textbf{AXPO}\,\scalebox{0.8}{\textit{(Ours)}}
            & 56.1\,\textsubscript{$\pm$1.4} & 79.0\,\textsubscript{$\pm$0.9} & 60.6\,\textsubscript{$\pm$0.8}
            & 87.8\,\textsubscript{$\pm$0.4} & 45.8\,\textsubscript{$\pm$1.1} & 83.2\,\textsubscript{$\pm$0.6} & 77.0\,\textsubscript{$\pm$4.3}
            & 25.9\,\textsubscript{$\pm$0.7} & 45.0\,\textsubscript{$\pm$0.4}
            & 62.3\,\textsubscript{$\pm$1.2} \\
        \midrule

        \multicolumn{11}{l}{\textit{Qwen3-VL-32B-Thinking} (inference-only baseline)} \\
        \midrule
        Base
            & 56.5\,\textsubscript{$\pm$1.4} & 83.3\,\textsubscript{$\pm$0.6} & 64.1\,\textsubscript{$\pm$1.1}
            & 89.1\,\textsubscript{$\pm$1.1} & 40.3\,\textsubscript{$\pm$2.4} & 85.2\,\textsubscript{$\pm$1.6} & 78.9\,\textsubscript{$\pm$3.7}
            & 22.8\,\textsubscript{$\pm$1.2} & 46.1\,\textsubscript{$\pm$1.6}
            & 62.9\,\textsubscript{$\pm$1.6} \\

        \bottomrule
    \end{tabular}
    }
\vspace{-0.2in}
\end{table*}

\autoref{tab:main-p1} reports Pass@1 averaged over four rollouts per question for compactness. To quantify rollout-induced variance, we re-evaluate every (model, benchmark) cell with \textbf{four independent evaluation rollouts} (each itself an average over four rollouts per question, matching the main-table protocol) and report the resulting mean\,$\pm$\,std in \autoref{tab:main-p1-std}. Per-cell std values fall in the $0.4$--$3.6$\,pp range for nearly all entries, with isolated outliers on the smallest (2B) and most variance-prone (HRBench-8K) settings. The column-averaged std for SFT\,+\,AXPO is $1.4$\,pp at 4B and $1.2$\,pp at 8B, both below the AXPO\,$-$\,SFT\,+\,GRPO Pass@1 deltas reported in \autoref{tab:main-p1} ($+1.4$ and $+1.8$\,pp respectively), so the headline gains lie above the per-rollout noise floor.

\subsection{Decomposing the SFT-then-RL Recipe}
\label{app:decomp-recipe}

\autoref{tab:main-p4} contains four pre-AXPO baselines per model size: Base (tools, no training), GRPO (RL from base), SFT (SFT only), and SFT\,+\,GRPO. We additionally report a Non-agent (no tools, no training) configuration in \autoref{tab:non-agent-vs-base} as a sanity check on the marginal value of inference-time tool access, which is orthogonal to the agentic-RL story. Reading the average Pass@1 / Pass@4 columns across these baselines isolates the contribution of each pipeline stage and motivates the SFT-then-RL recipe that AXPO builds on. All numbers in this section come from the \textbf{Average} columns of \autoref{tab:main-p1} (Pass@1) and \autoref{tab:main-p4} (Pass@4).
\begin{table*}[t]
    \caption{\textbf{Non-agent vs Base}: effect of turning tools on without any training, across nine multimodal benchmarks and four model sizes. \textit{Non-agent} disables tool calls (the model answers from internal reasoning alone). \textit{Base} enables the same tool interface as the trained methods but receives no post-training. Both Pass@1 and Pass@4 (\%, $\uparrow$) are reported. The Pass@1 / Pass@4 deltas isolate the marginal value of inference-time tool access at each scale, motivating the post-training pipeline analyzed in \appref{app:decomp-recipe}.}
    \vspace{-0.1in}
    \label{tab:non-agent-vs-base}
    \centering
    \scriptsize
    \resizebox{\textwidth}{!}{
    \begin{tabular}{l ccc cccc cc >{\columncolor{mygray}}c}
        \toprule
        \multicolumn{1}{c}{} &
        \multicolumn{3}{c}{\textbf{Reasoning}} &
        \multicolumn{4}{c}{\textbf{Perception}} &
        \multicolumn{2}{c}{\textbf{Search}} &
        \multicolumn{1}{c}{\multirow{2}{*}[-3pt]{\textbf{Avg.}}} \\
        \cmidrule(lr){2-4} \cmidrule(lr){5-8} \cmidrule(lr){9-10}
        \textbf{Method} &
        \shortstack{Math\\Vision} & \shortstack{Dyna\\Math} & \shortstack{Math\\-VR} &
        V$^{\star}$ & \shortstack{Visual\\Probe} & \shortstack{HRBen.\\-4K} & \shortstack{HRBen.\\-8K} &
        \shortstack{HR-MM\\Search} & \shortstack{MM\\Search} & \multicolumn{1}{c}{} \\
        \midrule

        \multicolumn{11}{l}{\textit{Pass@1}} \\
        \midrule
        \multicolumn{11}{l}{\textit{Qwen3-VL-2B-Thinking}} \\
        Non-agent
            & 27.9 & 66.5 & 24.8 & 64.8 & 18.2 & 57.8 & 57.3 & 5.1 & 9.5 & 36.9 \\
        Base
            & 18.6 & 60.5 & 16.9 & 63.0 & 5.0 & 64.6 & 55.9 & 8.2 & 27.7 & 35.6 \\
        \midrule
        \multicolumn{11}{l}{\textit{Qwen3-VL-4B-Thinking}} \\
        Non-agent
            & 44.8 & 76.1 & 45.9 & 79.6 & 21.9 & 74.4 & 68.1 & 6.1 & 13.3 & 47.8 \\
        Base
            & 48.0 & 76.6 & 50.8 & 80.6 & 24.8 & 73.6 & 67.5 & 16.9 & 37.8 & 53.0 \\
        \midrule
        \multicolumn{11}{l}{\textit{Qwen3-VL-8B-Thinking}} \\
        Non-agent
            & 52.5 & 75.5 & 50.6 & 78.0 & 27.6 & 70.8 & 68.1 & 8.5 & 15.3 & 49.7 \\
        Base
            & 47.1 & 75.9 & 54.9 & 77.7 & 31.8 & 72.8 & 66.1 & 21.0 & 42.7 & 54.4 \\
        \midrule
        \multicolumn{11}{l}{\textit{Qwen3-VL-32B-Thinking}} \\
        Non-agent
            & 61.2 & 84.2 & 64.2 & 84.4 & 24.8 & 81.4 & 74.4 & 11.2 & 19.8 & 56.2 \\
        Base
            & 56.5 & 83.3 & 64.1 & 89.1 & 40.3 & 85.2 & 78.9 & 22.8 & 46.1 & 62.9 \\
        \midrule

        \multicolumn{11}{l}{\textit{Pass@4}} \\
        \midrule
        \multicolumn{11}{l}{\textit{Qwen3-VL-2B-Thinking}} \\
        Non-agent
            & 43.8 & 79.4 & 39.4 & 77.5 & 31.1 & 73.0 & 75.5 & 10.2 & 16.3 & 49.6 \\
        Base
            & 32.6 & 76.6 & 28.9 & 81.7 & 15.1 & 79.0 & 77.0 & 18.7 & 46.3 & 50.7 \\
        \midrule
        \multicolumn{11}{l}{\textit{Qwen3-VL-4B-Thinking}} \\
        Non-agent
            & 62.0 & 82.6 & 60.4 & 89.0 & 31.1 & 83.5 & 78.5 & 9.8 & 20.3 & 57.5 \\
        Base
            & 64.8 & 82.6 & 65.1 & 95.8 & 44.3 & 83.5 & 81.0 & 27.5 & 55.7 & 66.7 \\
        \midrule
        \multicolumn{11}{l}{\textit{Qwen3-VL-8B-Thinking}} \\
        Non-agent
            & 70.7 & 85.0 & 63.7 & 85.3 & 38.7 & 79.0 & 77.0 & 16.4 & 22.7 & 59.8 \\
        Base
            & 63.5 & 85.2 & 68.3 & 92.7 & 55.7 & 85.5 & 82.5 & 33.8 & 59.0 & 69.6 \\
        \midrule
        \multicolumn{11}{l}{\textit{Qwen3-VL-32B-Thinking}} \\
        Non-agent
            & 78.9 & 91.2 & 75.3 & 91.6 & 33.0 & 88.5 & 80.5 & 16.4 & 30.0 & 65.0 \\
        Base
            & 73.0 & 89.4 & 73.9 & 97.4 & 64.2 & 90.5 & 90.0 & 38.0 & 59.3 & 75.1 \\

        \bottomrule
    \end{tabular}
    }
\vspace{-0.2in}
\end{table*}

\paragraph{Why is post-training needed at all?}
Comparing Non-agent and Base at the same model size measures the effect of \emph{turning tools on without any training}. The gain is strongly scale-dependent: Pass@1 / Pass@4 deltas are \dAgentTwoB at 2B, \dAgentFourB at 4B, \dAgentEightB at 8B, and \dAgentThirtyTwoB at 32B. At 2B the Base (agentic) baseline is actually \emph{worse} than the Non-agent baseline on Pass@1, showing that small models pay an agent-prompt overhead that outweighs the gain from correct tool use. Larger models recover Pass@1 and accumulate a Pass@4 ceiling gain, but even at 32B the gap to SFT\,+\,AXPO at 8B on Pass@4 (\eightBvsAgentPassFour) shows that inference-time tool access alone does not close the agentic gap. Post-training is therefore necessary at every trainable scale.

\paragraph{Why SFT before RL?}
Comparing GRPO (RL from base) against SFT\,+\,GRPO measures the value of an SFT initialization before RL. Pass@1 gaps shrink with size (\dSFTinitPassOne), but Pass@4 gaps remain positive at every scale (\dSFTinitPassFour). RL from the base policy can find a single high-reward mode (Pass@1 partially recovered at 8B), but it consistently sacrifices diversity, capping the Pass@4 ceiling~\citep{rlvr}.
SFT broadens the cold-start distribution over tool-using trajectories before group-based exploration begins. Without it, all-wrong tool-using subgroups dominate from step~1 and the policy collapses onto a narrow single-mode solution~\citep{deepeyesv2}.

\paragraph{Why RL after SFT?}
SFT minimizes log-likelihood on a fixed set of teacher trajectories: it can make the student \emph{imitate} the teacher's tool-use patterns, but the optimization signal lives entirely on the teacher's distribution. At test time the student samples from its own distribution and small token-level deviations from the teacher accumulate into prefixes the SFT objective never scored, so imitation alone does not guarantee that on-policy rollouts inherit the teacher's correctness (\emph{exposure bias}).
RL after SFT narrows this gap by training on the policy's own rollouts under an outcome reward, adapting the policy to its own induced trajectory distribution rather than to the teacher's~\citep{rlgeneralize}.
Empirically, comparing SFT against SFT\,+\,GRPO isolates this on-policy adaptation effect given the SFT initialization: Pass@1 improves at every scale (\dRLAfterSFTPassOne), the regime where on-policy reliability matters most, while Pass@4 gains are smaller and uneven (\dRLAfterSFTPassFour), consistent with GRPO consolidating mass on trajectories the SFT policy can already reach rather than discovering new ones. RL is therefore necessary on top of SFT, but standard GRPO's adaptation is biased toward the SFT manifold, motivating AXPO's \toolresample.

\paragraph{Why AXPO on top of SFT, instead of GRPO?}
The previous two paragraphs leave a structural gap: SFT sets the Pass@4 ceiling and RL sharpens within it, so further gains require a mechanism that simultaneously extends the ceiling \emph{and} sharpens. AXPO does both. Pass@1 gains over SFT\,+\,GRPO are \dPassOneTwoB / \dPassOneFourB / \dPassOneEightB $\ $at 2B/4B/8B, and Pass@4 gains are \dPassFourTwoB / \dPassFourFourB / \dPassFourEightB. The Pass@4 lift is the diagnostic signature that AXPO operates on a different axis from SFT-then-GRPO sharpening: by recovering all-wrong tool-using subgroups whose gradient signal would otherwise vanish (\autoref{sec:gap}), AXPO turns previously dead continuations into trainable ones, expanding the policy's reachable correct-trajectory set rather than only redistributing mass within it.

\subsection{Generalization to an Unseen Tool: Image Search}
\label{app:image-search-gen}
\begin{table}[t]
    \caption{\textbf{Generalization to an unseen tool.} MMSearch Pass@1 / Pass@4 (\%, $\uparrow$), with and without an image-search tool added at inference time. Image search is not in the training tool set (Python, Tavily web search, image zoom-in). AXPO converts the new tool into Pass@1 gains more effectively than GRPO ($+4.6$\,pp vs.\ $+2.8$\,pp), while Pass@4 gains are similar across the two methods.}
    \label{tab:image-search-gen}
    \centering
    \small
    \begin{tabular}{l l c c}
        \toprule
        & & SFT\,+\,GRPO & SFT\,+\,AXPO \\
        \midrule
        \multirow{2}{*}{w/o image search}
            & Pass@1 & 44.0 & \bf 45.0 \\
            & Pass@4 & 59.3 & \bf 61.0 \\
        \midrule
        \multirow{2}{*}{w/\phantom{o} image search}
            & Pass@1 & 46.8 & \textbf{49.6} \\
            & Pass@4 & 65.3 & \textbf{66.7} \\
        \bottomrule
    \end{tabular}
\end{table}

Training in \autoref{sec:setup} uses three tool families: a Python interpreter, a Tavily web search engine, and an image zoom-in tool. None of these is an \emph{image-based} retrieval tool that takes an image as the query and returns visually similar pages, which we cannot directly utilize due to excessive API costs.
To probe whether AXPO's tool-using policy transfers beyond the training tool inventory, we extend the inference-time tool set with an image-search tool and re-evaluate on \textbf{MMSearch}~\citep{mmsearch}, the search benchmark with the largest fraction of image-grounded queries. Since we did not have access to a production image-retrieval endpoint at the budget required for full evaluation, we approximate the tool by prompting GPT-5.4 with the query image and instructing it to return the kind of structured response a visual-search API would emit (caption, candidate entity names, related URLs). This proxy is imperfect, but it preserves the key property the experiment probes: an entirely new tool interface, unseen at training time, whose output the model must integrate into its reasoning.

\autoref{tab:image-search-gen} shows the result. \emph{Without} image search at inference, SFT\,+\,GRPO and SFT\,+\,AXPO are essentially tied (Pass@1: $44.0$ vs.\ $45.0$; Pass@4: $59.3$ vs.\ $61.0$). \emph{With} image search at inference, the gap opens on Pass@1: AXPO gains $+4.6$\,pp ($45.0 \to 49.6$) while GRPO gains only $+2.8$\,pp ($44.0 \to 46.8$); on Pass@4 both methods gain similarly ($+6.0$ and $+5.7$). The Pass@1 asymmetry suggests AXPO not only \emph{can} use the unseen tool when given enough samples (Pass@4 captures this) but converts that capability into reliable first-shot answers, whereas GRPO's tool-using policy is more narrowly fitted to the training tool inventory. We read this as evidence that resampling-induced exposure to diverse tool-call resolutions during training transfers to tool families the model has never been trained on, although the claim is bounded to one benchmark and one new tool type.

\subsection{Detailed Ablation Analysis}
\label{app:ablation-details}

\autoref{tab:ablation} ablates each AXPO design choice on 8B at matched compute. All numbers below are 8B Pass@1 averages on the four-benchmark composition (MathVision, V$^{\star}$, VisualProbe, HR-MMSearch).

\paragraph{Prefix fixing.}
Removing prefix fixing reduces AXPO to $52.1$, near the SFT\,+\,GRPO baseline of $51.9$: each resample becomes a fresh rollout from $\pi_\theta(\cdot \mid \vx)$, so only a $q$-fraction of resamples are tool-using and the $(1{-}q)$ waste factor that \autoref{prop:branching} removes returns in full.

\paragraph{Tool-using resampling target.}
Targeting questions where all no-tool rollouts fail for resampling instead reaches $51.7$, since the policy already covers the no-tool subgroup under GRPO and additional samples there cannot recover the missing tool-using signal. Resampling pays off only when the resampling target is the under-covered tool-using subgroup.

\paragraph{Uncertainty-based prefix selection.}
Replacing uncertainty rank with random ranking drops to $51.0$ ($-2.9$\,pp): the prefix is still fixed so the coverage gain is preserved, but random ranking routes the budget to prefixes where the policy has already converged toward a wrong continuation, wasting resamples on determined commitments.

\paragraph{Prefix credit.}
Dropping the recovery-indicator credit on the source prefix yields $51.4$ ($-2.5$\,pp): prefix tokens receive no positive signal even when resampling recovers a continuation, so the policy is not steered toward producing high-yield prefixes in the first place.

\paragraph{Per-prefix advantage group.}
Merging resampled continuations into the source group instead of forming a per-prefix advantage yields $51.2$ ($-2.7$\,pp): the shared prefix then collects contradictory advantages from the failed source and successful resamples while no-tool successes in the source group dilute the recovered signal.

\subsection{Detailed Comparison Analysis}
\label{app:comparison-details}

\autoref{tab:comparison} compares AXPO to alternative training recipes. All numbers are 8B Pass@1 averages.

\paragraph{Reward shaping does not recover the gain of AXPO.}
A tool penalty~\citep{otc} drops to $46.0$ and a tool bonus~\citep{pyvision-rl, pixelreasoner, deepeyes} reaches only $50.8$, both \emph{below} the GRPO baseline of $51.9$. Reward-channel interventions either suppress tool use globally (penalty) or reward tool calls regardless of whether they help (bonus); neither addresses the coverage problem that AXPO targets, which is conditional on the prefix being tool-using and on resampling restoring at least one correct continuation.

\paragraph{Extra rollout budget does not close the gap.}
Doubling the GRPO rollout count ($\times 4$ exceeding AXPO's extra compute) reaches $52.4$, still $1.5$\,pp below SFT\,+\,AXPO. The extra budget re-derives non-tool-using prefixes and cannot break the all-wrong tool-using subgroups that resampling is designed to recover; throwing more samples at the same $(1{-}q)$ waste factor does not change the structural problem.

\paragraph{Alternative RL algorithms.}
We replace GRPO with three recently proposed alternatives. RLTF~\citep{rltf}, an algorithm that utilizes the text feedback from the external teacher, reaches only $43.3$ Avg with $40\%$ tool utilization, suffering tool-call collapse over training (consistent with reports of hint leakage degrading text-feedback RL on agentic tasks). CISPO~\citep{cispo}, an importance-sampling variant, reaches $47.8$ Avg: its clip rule reduces gradient bias but does not address the all-wrong subgroup whose advantage is zero by construction. ARPO~\citep{arpo}, an agent-specific branching method that draws additional continuations after a tool observation when post-observation entropy is high, is the closest comparison and the strongest competitor at $51.3$ Avg, yet still trails AXPO by $2.6$\,pp: post-observation branching cannot recover when the tool call itself is the failure mode, which is the regime \autoref{sec:gap} identifies as binding. The RL-algorithm sweep confirms AXPO's gain is not a generic effect of any agent-aware RL trick but specifically reflects \toolresample's coverage restoration on tool-using thinking prefixes.

\subsection{Comparison with Prior Agentic VLMs}
\label{app:prior-agentic-vlms}
\begin{table}[t]
    \caption{Comparison with prior agentic VLM systems on five benchmarks where numbers are publicly available. All numbers are Pass@1 (\%, $\uparrow$). ``--'' indicates the method does not report on that benchmark. Average is computed only over methods with all five entries. Best per column is \textbf{bolded}; second-best is \underline{underlined}. Prior agentic systems use different base models (mostly Qwen2.5-VL-7B), so absolute values are not strictly comparable; the \textit{$\Delta$ vs.\ base} rows report each group's best improvement over its own base, isolating the post-training gain from the base-model gap. AXPO delivers a larger average lift over its base ($+8.7$\,pp) than the strongest prior agentic recipe over Qwen2.5-VL-7B ($+6.9$\,pp).}
    \label{tab:model-comp}
    \centering
    \small
    \resizebox{\textwidth}{!}{
    \begin{tabular}{l ccccc >{\columncolor{mygray}}c}
        \toprule
        \textbf{Method} & MathVision & DynaMath & V$^{\star}$ & HR-Bench-4K & HR-Bench-8K & \multicolumn{1}{c}{\textbf{Avg.}} \\
        \midrule
        \multicolumn{7}{l}{\textit{Qwen2.5-VL-7B base + prior agentic methods}} \\
        Qwen2.5-VL-7B-Instruct~\citep{qwen2.5-vl}
            & 25.6 & 53.3 & 78.5 & 71.6 & 67.9 & 59.4 \\
        PixelReasoner~\citep{pixelreasoner}
            & -- & -- & 84.3 & 74.0 & 66.9 & -- \\
        Mini-o3~\citep{mini-o3}
            & -- & -- & \underline{88.2} & 77.5 & 73.3 & -- \\
        DeepEyes~\citep{deepeyes}
            & 26.6 & 55.0 & 85.6 & 75.1 & 72.6 & 63.0 \\
        Thyme~\citep{thyme}
            & 27.6 & -- & 82.2 & 77.0 & 72.0 & -- \\
        DeepEyes-v2~\citep{deepeyesv2}
            & 28.9 & 57.2 & 81.8 & \underline{77.9} & \underline{73.8} & -- \\
        PyVision-RL~\citep{pyvision-rl}
            & 28.7 & 61.6 & \textbf{88.7} & 78.1 & 74.3 & 66.3 \\
        \rowcolor{mygray!50}
        \textit{\;\;$\Delta$ vs.\ base (best prior)}
            & \textit{+3.3} & \textbf{\textit{+8.3}} & \textbf{\textit{+10.2}} & \textit{+6.5} & \textit{+6.4} & \textit{+6.9} \\
        \midrule
        \multicolumn{7}{l}{\textit{Qwen3-VL-8B-Thinking base + ours}} \\
        Qwen3-VL-8B-Thinking (Agent)~\citep{Qwen3VL}
            & \underline{47.1} & \underline{75.9} & 77.7 & 72.8 & 66.1 & \underline{67.9} \\
        \rowcolor{myblue}
        \textbf{SFT\,+\,AXPO}\,\scalebox{0.8}{\textit{(Ours)}}
            & \textbf{56.1} & \textbf{79.0} & 87.8 & \textbf{83.3} & \textbf{77.0} & \textbf{76.6} \\
        \rowcolor{myblue!50}
        \textit{\;\;$\Delta$ vs.\ base (ours)}
            & \textbf{\textit{+9.0}} & \textit{+3.1} & \textit{+10.1} & \textbf{\textit{+10.5}} & \textbf{\textit{+10.9}} & \textbf{\textit{+8.7}} \\
        \bottomrule
    \end{tabular}
    }
\end{table}

\autoref{tab:model-comp} situates SFT\,+\,AXPO against prior agentic VLM systems on the five benchmarks for which their published numbers are available (MathVision, DynaMath, V$^{\star}$, HR-Bench-4K, HR-Bench-8K) from previous work~\citep{pyvision-rl, deepeyesv2}.
This comparison is not strictly controlled: prior systems use different base models (mostly Qwen2.5-VL-7B), different training dataset scale, and report on benchmark subsets, so part of any gap reflects base-model improvements rather than method choice. 
With that caveat, SFT\,+\,AXPO at 8B leads on four of the five benchmarks and on the cross-benchmark average ($76.6$ vs.\ $66.3$ for the strongest prior system reporting on all five). The exception is V$^{\star}$, where pixel-level reasoning systems (Mini-o3, PyVision-RL) sit within $\sim$1\,pp; this is a benchmark where image zoom-in is the load-bearing tool and tool-using rollouts are already near saturation across systems, so the headroom for gradient-rescue is small. 
The largest gaps appear on math-over-image benchmarks (MathVision, DynaMath), where prior agentic VLM training has invested less and where AXPO's tool-call gradient survival is most directly load-bearing.

\subsection{Confidence as a Tractable Proxy for Tool-Call Entropy}
\label{app:conf-vs-entropy}

\autoref{sec:method} ranks candidate prefixes by the mean probability the source policy assigned to the tool-call tokens of $\tau^{\text{src}}$ (henceforth \emph{confidence}). The natural alternative is the predictive entropy $H(\pi_\theta(\cdot \mid \vx, \vy_{<t}))$ of the next-token distribution at each tool-call position, but this choice is precluded by training-loop structure. Per-token sampling probabilities are emitted essentially for free by the rollout engine (vLLM~\citep{vllm}), whereas predictive entropy requires either an extra forward pass over the full vocabulary under the trained policy (FSDP) or vLLM-side full-vocabulary logprob expansion at decode time. Our RL pipeline under verl~\citep{verl} alternates rollout (vLLM) and gradient updates (FSDP), so inserting a per-prefix entropy computation would force an additional vLLM\,$\to$\,FSDP\,$\to$\,vLLM round-trip per training step (or equivalently expensive vLLM-side full-vocabulary logprob expansion), a cost we wished to avoid.

\begin{figure}[t]
    \centering
    \includegraphics[width=0.65\linewidth]{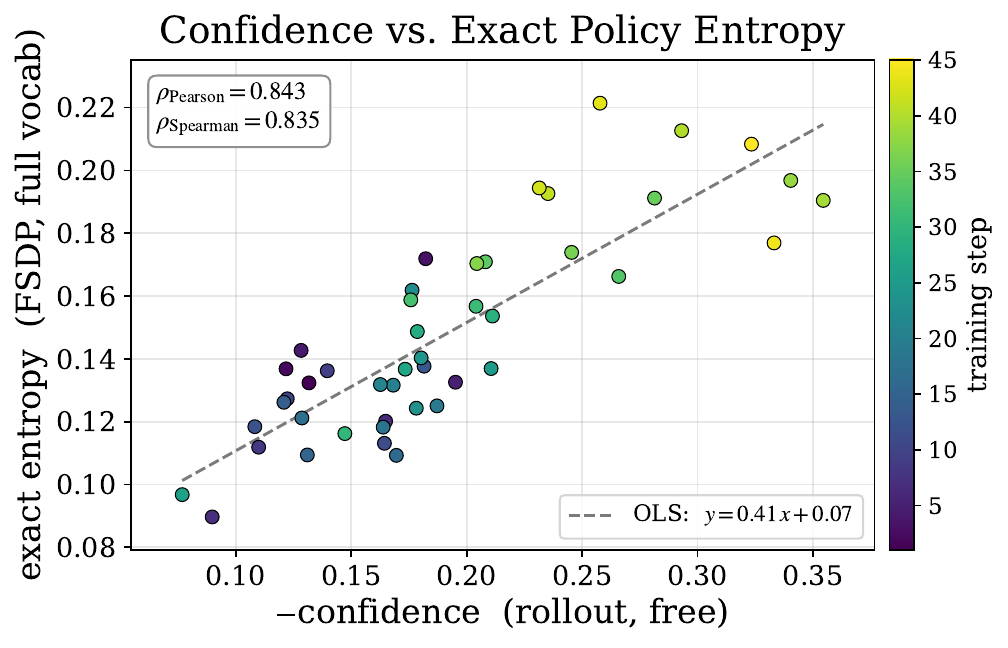}
    \caption{\textbf{Confidence vs.\ exact policy entropy at tool-call positions.} For each training step, we plot the mean confidence (vLLM rollout, free) against the exact policy entropy (FSDP forward, full vocabulary) on tool-call tokens of failed tool-using rollouts. The two signals are strongly correlated (Pearson $\rho = 0.843$, Spearman $\rho = 0.835$), justifying confidence as a tractable uncertainty proxy.}
    \label{fig:conf-vs-entropy}
\end{figure}

To verify that confidence preserves the ranking an entropy-based scheme would induce, we measured the per-step correlation between mean-token-confidence and exact policy entropy on tool-call positions of failed tool-using rollouts collected during training (\autoref{fig:conf-vs-entropy}). Across training steps, the two signals are strongly correlated (Pearson $\rho = 0.843$, Spearman $\rho = 0.835$, OLS fit $y = 0.41 x + 0.07$ on entropy vs negated confidence). Confidence therefore preserves the relative ordering an entropy-based scheme would produce, and we adopt it as the ranking signal in AXPO.

\section{Extended Related Works}
\label{app:related-works}

This appendix expands on \autoref{sec:related} with the longer descriptions that the main text abridges.

\paragraph{Two-stage SFT-then-RL recipe for agentic VLMs.}
Recent agentic VLM systems share a two-stage recipe: supervised fine-tuning on tool-using trajectories distilled from larger teachers~\citep{FireAct, AgentDistill}, followed by outcome-reward RL~\citep{pyvision-rl, deepeyesv2, deepeyes}. They differ mainly in the distillation corpus, the tool interface, and the choice of base model. Most prior agentic-RL VLM studies focus on a single tool family (typically image manipulation), whereas our setup combines code, search, and image zoom-in on a thinking-capable base (Qwen3-VL-Thinking)~\citep{Qwen3VL}, a configuration under which the RL stage becomes the primary bottleneck.

\paragraph{Group-relative RL and BroRL.}
RL with verifiable outcome rewards is the standard post-SFT stage for reasoning models~\citep{r1, o1}. GRPO~\citep{GRPO, DAPO, prorl} samples $N$ rollouts per question, normalizes rewards within the group, and assigns the resulting advantage uniformly to every token. BroRL~\citep{brorl} argues that increasing $N$ reduces unsampled blind spots at a rate determined by per-token baseline probability. Both analyses were developed for thinking-only trajectories, where the rollout structure is uniform across tokens and the per-token baseline probability decays slowly. \autoref{prop:branching} shows that AXPO's prefix-fixed resampling dominates raw scaling in the regime where tool-call tokens sit at low policy probability, exactly the failure mode \autoref{sec:gap} diagnoses.

\paragraph{Mid-trajectory branching for tool-using RL.}
ARPO~\citep{arpo} and AEPO~\citep{aepo} adapt group-relative RL to tool-using rollouts by branching \emph{after} a tool observation, drawing additional continuations from $\pi_\theta(\cdot \mid \vx, \vt_1, \va_1, \vo_1)$ when post-observation entropy is high. The exploration target is the continuation given a fixed tool call, which absorbs variance introduced by the tool output but cannot recover when the tool call itself is wrong (wrong sub-image, wrong python code, wrong search query). AXPO instead branches \emph{at} the tool-call boundary and triggers only on all-wrong tool-using subgroups, with a per-prefix advantage that credits the source prefix exactly when its branched continuation succeeds, making it a targeted intervention on the regime \autoref{sec:gap} identifies as binding rather than a general rollout-budget reshaper. Empirically, AXPO outperforms ARPO on the 8B comparison ($53.9$ vs.\ $51.3$ Avg, \autoref{tab:comparison}), consistent with the diagnosis that the tool call itself, not the post-observation continuation, is the dominant failure locus. The two strategies are compatible, and combining tool-call and post-observation branching is a natural extension we leave to future work.

\section{Coverage Argument for \Toolresample}
\label{app:coverage}

This appendix formalizes \autoref{prop:branching}: prefix-fixed resampling from a tool-committed prefix dominates raw sampling at recovering correct tool-using rollouts whenever the selected prefix satisfies $p(\vt_1) \geq q\, p^{\text{tool}}$, with strict inequality once $p(\vt_1) > q\, p^{\text{tool}}$. The mechanism is $q \to 1$ by construction: every resampled continuation from a tool-committed prefix is tool-using, so raw's $(1-q)$ waste factor vanishes and the threshold a prefix must clear sits at $q\, p^{\text{tool}}$, strictly below the average per-tool-using-rollout success $p^{\text{tool}}$.

\paragraph{Setup.}
Let $\pi_\theta$ be the policy, $\vx$ a question, and $R(\tau) \in \{0,1\}$ the outcome reward. Decompose $\tau = (\vt_1, \vy_{\geq 1})$ into a thinking prefix and its continuation (tool call, tool output, and the remainder of the trajectory). Let $\mathcal{T} = \{\vt_1 : \vt_1 \text{ includes the opening } \toolcall \text{ tag}\}$ be the set of tool-committed prefixes, and define
\[
    q := \Pr_{\pi_\theta}[\vt_1 \in \mathcal{T} \mid \vx], \qquad
    p(\vt_1) := \Pr_{\pi_\theta}[R=1 \mid \vx, \vt_1], \qquad
    p^{\text{tool}} := \mathbb{E}_{\vt_1 \sim \pi_\theta(\cdot \mid \vx,\, \vt_1 \in \mathcal{T})}[p(\vt_1)].
\]
So $q$ is the rate at which the policy commits to tool use from $\vx$, and $p^{\text{tool}}$ is the per-tool-using-rollout success rate. Both are directly measurable (\autoref{fig:analysis-a}, \autoref{fig:analysis-b}). We compare two samplers, each drawing $N$ trajectories, on the event \emph{``at least one sample is a correct tool-using rollout''}:
\begin{itemize}[leftmargin=*, topsep=2pt, itemsep=1pt]
    \item \textbf{Raw rollout}: $\tau \sim \pi_\theta(\cdot \mid \vx)$. Per-sample probability of a correct tool-using rollout is $\Pr[\vt_1 \in \mathcal{T}] \cdot \Pr[R=1 \mid \vt_1 \in \mathcal{T}] = q\, p^{\text{tool}}$.
    \item \textbf{Prefix-fixed resampling}: fix $\vt_1^{\text{src}} \in \mathcal{T}$ and draw continuations from $\pi_\theta(\cdot \mid \vx, \vt_1^{\text{src}})$. Every resampled continuation is tool-using by construction ($q \to 1$), and per-resample success is $p(\vt_1^{\text{src}})$.
\end{itemize}
Coverage, i.e., the probability that at least one of the $N$ samples is a correct tool-using rollout, equals $1 - (1-p_{\text{eff}})^N$ with $p_{\text{eff}} = q\, p^{\text{tool}}$ under raw rollout and $p_{\text{eff}} = p(\vt_1^{\text{src}})$ under resampling.

\paragraph{Proof of \autoref{prop:branching}.}
The function $f(p) = 1 - (1-p)^N$ has derivative $f'(p) = N(1-p)^{N-1}$, non-negative on $[0,1]$ and strictly positive on $[0,1)$, so $f$ is non-decreasing on $[0,1]$ and strictly increasing on $[0,1)$. Under the assumption $p(\vt_1^{\text{src}}) \geq q\, p^{\text{tool}}$, applying $f$ to both sides gives $1 - (1-p(\vt_1^{\text{src}}))^N \geq 1 - (1 - q\, p^{\text{tool}})^N$. Strictness follows whenever $p(\vt_1^{\text{src}}) > q\, p^{\text{tool}}$ and $q\, p^{\text{tool}} \in (0,1)$. The mechanism is $q \to 1$ by construction: resampling eliminates raw's wasted $(1-q)$ fraction of non-tool rollouts, so the baseline threshold is $q\, p^{\text{tool}}$ rather than $p^{\text{tool}}$.

\paragraph{Consequence for GRPO training.}
Under GRPO with binary rewards, coverage on correct tool-using rollouts determines whether tool-call tokens receive a \emph{positive} (reinforcing) gradient. If the tool-using subgroup contains no correct rollout, the tool-call tokens get no positive signal: the subgroup is either empty (no tool-call tokens at all in the group) or all-wrong with non-positive group-normalized advantage --- zero in fully all-wrong groups ($r_i - \mathrm{mean}(\{r_j\}) = 0$) and \emph{negative} in mixed-success groups where some no-tool rollouts succeed (which actively penalizes tool use). Hence $1 - (1-p_{\text{eff}})^N$ is exactly the probability that a group produces a positive (reinforcing) learning signal at the tool call. \autoref{prop:branching} then says that prefix-fixed resampling strictly breaks the all-wrong tool-using subgroups identified in \autoref{fig:analysis-b} at any fixed $N$.

\paragraph{Satisfying the threshold.}
By the definition of $p^{\text{tool}}$ as the conditional expectation over the tool-committed prefix distribution $\pi_\theta(\cdot \mid \vx,\, \vt_1 \in \mathcal{T})$, $\mathbb{E}_{\vt_1}[p(\vt_1)] = p^{\text{tool}} > q\, p^{\text{tool}}$ whenever $q < 1$, so the expected prefix lies strictly above the \autoref{prop:branching} threshold. All prefixes at or above the conditional mean satisfy the assumption, and so do below-average prefixes down to $q\, p^{\text{tool}}$; only prefixes with strongly subpar per-branch success ($p(\vt_1) < q\, p^{\text{tool}}$) can underperform raw sampling. Empirically, $p(\vt_1)$ is far from constant across $\mathcal{T}$: fixing a failed tool-using rollout's prefix and resampling its tool call recovers $\sim$15\% of originally-failed tool-using subgroups.

\paragraph{Relation to BroRL.}
Our target is tool use, and the diagnostics in \autoref{sec:gap} establish exactly the regime where naive rollout scaling fails to help: tool use is rare (\autoref{fig:analysis-a}: $q \in [0.2, 0.35]$) and all-wrong rates on tool-using subgroups are high (\autoref{fig:analysis-b}), so tool-call tokens sit in a low-probability regime under the policy. \citet{brorl} show that scaling $N$ reduces unsampled blind spots at rate $p^2(1-p)^N$ (Theorem 1, Lemma 2); this decay is per-token, fast for high-$p$ regions and slow for low-$p$ regions. Raw scaling therefore preferentially fills the thinking-only blind spot while leaving the tool-call blind spot under-sampled. \autoref{prop:branching} sidesteps this by relocating samples: conditioning on a tool-committed prefix eliminates the $(1-q)$ waste factor so the local per-branch success becomes $p(\vt_1^{\text{src}})$, and the same budget $N$ reaches the tool region effectively. BroRL's guarantee is uniform in token type; AXPO targets the specific region that the low tool-use attempt rate has made hardest to cover.

\section{Limitations and Future Work}
\label{app:limitations}

We expand on the limitations sketched in the main-text conclusion and discuss future directions.

\paragraph{Verifiable outcome rewards.}
AXPO's subgroup-level trigger (``all-wrong tool-using subgroup'') and per-prefix advantage in Eq.~\ref{eq:axpo-adv} both rely on a binary, automatically verifiable outcome signal $r \in \{0,1\}$. Tasks where verifiability is partial (open-ended generation~\citep{NEURIPS2025_e5849736, lee2025maskingteacherreinforcingstudent}, multi-criterion subjective evaluation) require a different definition of ``failed subgroup'' before \toolresample applies. The resampling mechanism itself is reward-agnostic, but the trigger and the all-wrong learning signal argument in \autoref{sec:gap} are not.

\paragraph{Compute scope of training.}
Trained models are at most 8B parameters; 32B is reported inference-only as the SFT teacher. This boundary reflects available compute rather than a methodological choice~\citep{kim2026visualtokenpruningfails, lee2025maskingteacherreinforcingstudent, yu2026hideseereasoningprefixmasking, lee2026genrecalgenerationrecalibrationlarge, Lee_2025_CVPR}.
Whether the Thinking-Acting Gap and AXPO's recovery generalize to 32B+ trainable scale, where the base policy is closer to the tool-use ceiling and the all-wrong-subgroup fraction may be smaller, is open.
Furthermore, although the teacher model can theoretically scale to 100B+ parameters, analyzing the effects of teacher scaling on downstream reinforcement learning (RL) remains an area for future work.

\paragraph{Future work: richer tool environments.}
Our tool inventory covers three families that span numerical, retrieval, and visual modalities (Python interpreter, web search, image zoom-in), each shown to be load-bearing in prior agentic VLM work~\citep{deepeyesv2, pyvision-rl, mini-o3}; this multi-tool setup is already broader than most prior agentic-VLM RL studies, which typically focus on a single tool family. One notable absence is an image-search tool, which prior works~\citep{deepeyesv2,hrmmsearch} report results with but which we could not include during training and full evaluation due to per-call API costs at our scale (we instead approximate it via a GPT-5.4 prompt for the unseen-tool study in \appref{app:image-search-gen}). Tools with longer execution latency (browser agents, GUI control, terminal sessions) and tools with high per-call cost (LLM-as-tool)~\citep{toolorchestra} introduce additional considerations that we do not address.
Extending AXPO to such richer~\citep{lee2025refinebenchevaluatingrefinementcapability, Lee_2025_ICCV, yu2024sparkmultivisionsensorperception}, longer-horizon environments is a natural next step.

\paragraph{Future work: generalization to other base models.}
All experiments use Qwen3-VL-Thinking, which is currently the strongest open thinking-capable VLM with reliable tool-using ability and the model best supported by our RL framework. The Thinking-Acting Gap is a property of RL on an agentic reasoning model rather than of any specific architecture~\citep{NEURIPS2024_473a9a75, 10.1007/978-3-031-72967-6_16, lee-etal-2024-collavo, lee2024phantomlatentlargelanguage}, and we expect the diagnostics and AXPO's coverage gain to transfer; replicating on the most recent Qwen3.5 models~\citep{qwen35}, once RL framework support matures, would help establish the gap as a general phenomenon and rule out base-specific confounds.

\section{Broader Impacts}
\label{app:broader-impacts}

AXPO is a methodological improvement to the RL training of agentic VLMs and is additive on top of a publicly released base model (Qwen3-VL-Thinking~\citep{Qwen3VL}), so it does not introduce a new capability class. Among the three tool families used in this work, the Python interpreter raises standard security concerns for code-executing agents. We can run all tool calls in a sandboxed execution environment during both training and evaluation, consistent with prior agentic-VLM RL work~\citep{pyvision}. Beyond this sandboxing, broader safety considerations~\citep{NEURIPS2021_8e5e15c4,Lee_2022_CVPR, Kim_2023_CVPR, Lee_2023_ICCV, lee2021towards, 10222502, KIM2026112173} (e.g., misuse of tool-grounded outputs, causality, robustness, deployment-time guardrails, downstream effects of cheaper agentic deployments) are out of scope for this work and were not addressed.

\newpage
\section{Qualitative Examples: GRPO vs.\ AXPO}
\label{app:qual-examples}

We show three side-by-side trajectories where SFT\,+\,AXPO succeeds (4/4) and SFT\,+\,GRPO fails (0/4) at the 8B scale, one per modality (perception, reasoning, search). For each example we display the question, the correct answer, and one representative trajectory per method. Thoughts are extensively abbreviated with ellipses for brevity; tool calls and outputs are reproduced verbatim.

\subsection{Perception: tiny-text reading on a Hong Kong street scene}
\label{app:qual-perception}

\begin{questionbox}
\textbf{Benchmark}: VisualProbe-hard~\citep{mini-o3}. \textbf{Ground truth}: \texttt{2563}.

\textbf{Prompt}: ``What is the number to the left of 1000?''
\begin{center}
\includegraphics[width=0.55\linewidth]{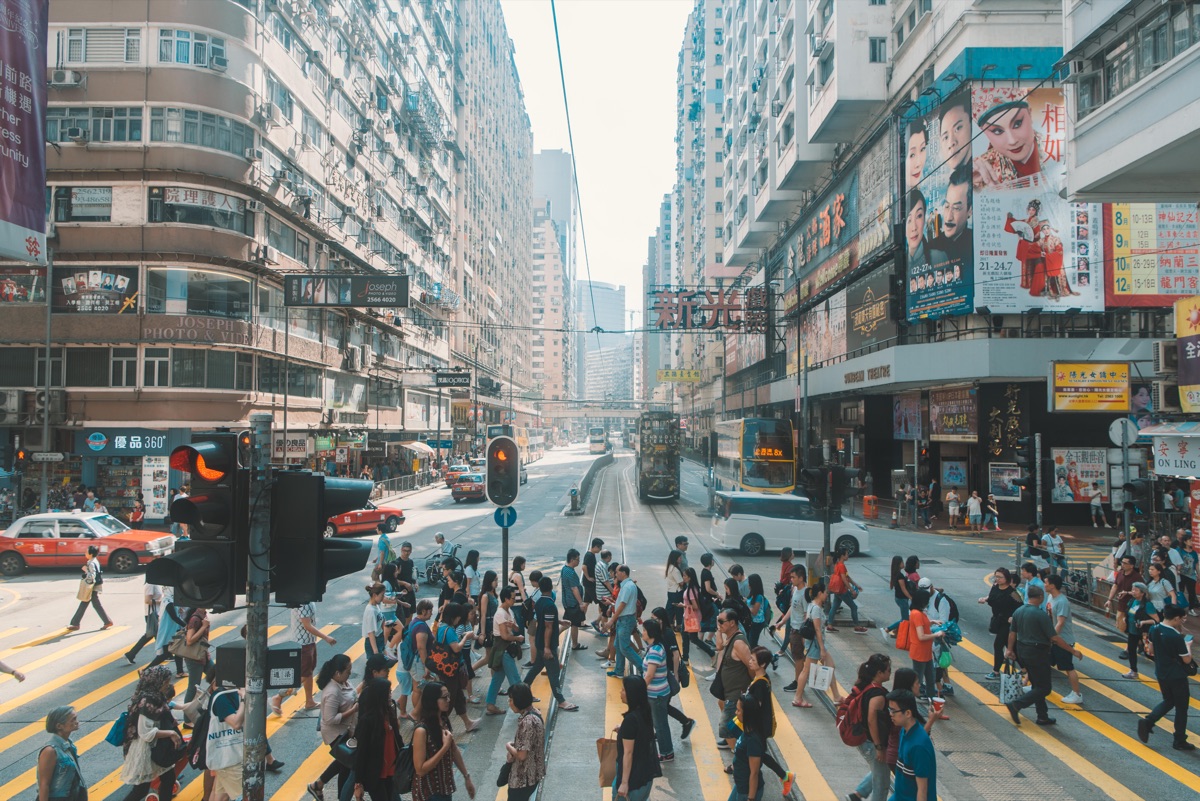}
\end{center}
\end{questionbox}

\begin{axpobox}
\textbf{Step 0, \think.} \emph{``\dots checking phone numbers and license plates\dots looking at the yellow sign on the right side under SUNBEAM THEATRE, there's a section with `Tel: 2563 1000'\dots Let me confirm with image\_zoom\_in.''}

\textbf{Step 0, \toolcall.}\\
{\small\ttfamily image\_zoom\_in(\{"bbox\_2d": [870, 450, 945, 517], "label": "Telephone number sign", "img\_idx": 0\})}
\begin{center}
\includegraphics[width=0.22\linewidth]{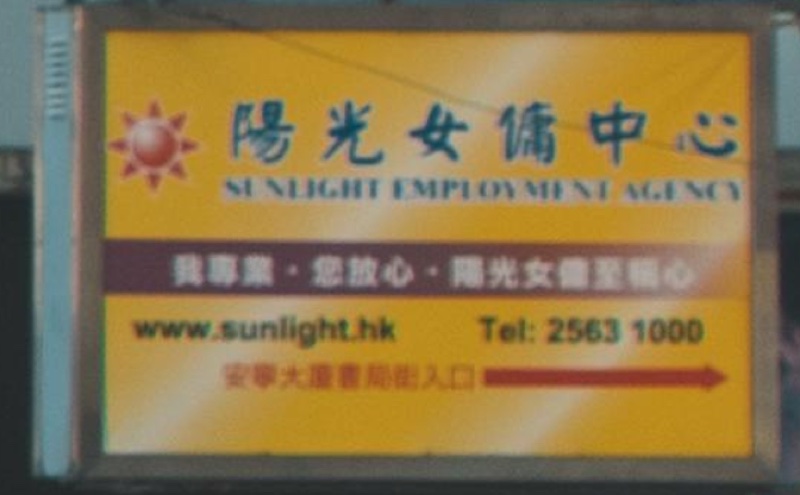}
\end{center}

\textbf{Step 1, \think.} \emph{``The cropped image clearly shows `Tel: 2563 1000'. The number to the left of 1000 is 2563.''}

\textbf{\answer} \texttt{2563} \(\checkmark\)
\end{axpobox}

\begin{grpobox}
\textbf{Step 0, \think.} \emph{``\dots checking traffic lights\dots license plates\dots bus route numbers\dots none match 1000\dots maybe the question is a trick. The number to the left of 1000 in numerical order is 999\dots Let me try image\_zoom\_in\dots [reasoning continues without ever emitting a tool call]\dots Given that this is a common math question, the answer is likely 999.''}

\textbf{No tool calls were emitted.}

\textbf{\answer} \texttt{999} \(\times\)
\end{grpobox}

The thinking prefix in both rollouts identifies the right candidate region, but only AXPO commits to a tool call; GRPO talks itself into a numerical-order guess (\texttt{999}) and never invokes \texttt{image\_zoom\_in}, despite repeatedly mentioning it as a candidate action. All four GRPO rollouts on this question follow the same pattern.

\subsection{Reasoning: geometry disambiguation via python visualization}
\label{app:qual-reasoning}

\begin{questionbox}
\textbf{Benchmark}: MathVision~\citep{mathvision}. \textbf{Ground truth}: \texttt{8}.

\textbf{Prompt}: ``Three semi-circles as shown. $ABEF$ is a rectangle and the radius of each of the semi-circles is 2 cm. $E$ and $F$ are the centers of the bottom semi-circles. The area of the shaded region (in cm$^2$) is:''
\begin{center}
\includegraphics[width=0.32\linewidth]{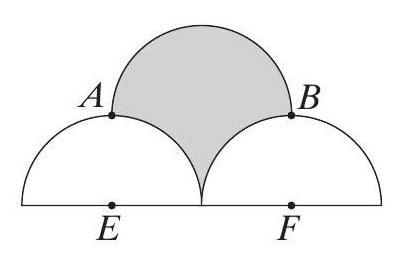}
\end{center}
\end{questionbox}

\begin{axpobox}
\textbf{Step 0, \think.} \emph{``\dots Setting $E=(0,0)$, $F=(4,0)$, the top semicircle has center $(2,2)$ and radius $2$. Computing sector minus triangle for each side gives an overlap of $\pi-2$\dots Shaded $= 2\pi - 2(\pi-2) = 4$. The result matches the area of a $2\times 2$ square, suggesting correctness. Let me also visualise the configuration to confirm.''}

\textbf{Step 0, \toolcall.}\\
{\small\ttfamily python(\{"code": "import matplotlib.pyplot as plt; import numpy as np;\dots ax.plot(x1, np.sqrt(4-x1**2), 'b'); ax.plot(x2, np.sqrt(4-(x2-4)**2), 'b'); ax.plot(x3, 2-np.sqrt(4-(x3-2)**2), 'r'); ax.add\_patch(plt.Polygon([(0,2),(0,0),(4,0),(4,2)], color='gray', alpha=0.5));\dots"\})}

\textbf{Tool output (matplotlib figure).}
\begin{center}
\includegraphics[width=0.45\linewidth]{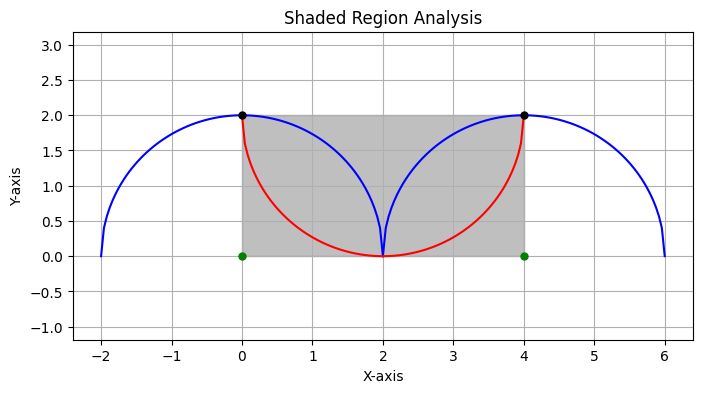}
\end{center}

\textbf{Step 1, \think.} \emph{``\dots The plot shows the bottom semicircles curving \textbf{upward} from $EF$ to $A$ and $B$ (forming quarter-disks at corners $E,F$, total area $2\pi$) and the top semicircle curving \textbf{downward} from $AB$ to $(2,0)$ (a half-disk of area $2\pi$). Since the two areas match, by area-swap the shaded region rearranges to the full rectangle $ABEF$: $4 \times 2 = 8$.''}

\textbf{\answer} \texttt{8} \(\checkmark\)
\end{axpobox}

\begin{grpobox}
\textbf{Step 0, \think.} \emph{``\dots Top semicircle area $= 2\pi$. Each side gives a sector of $\pi/2$ minus a right triangle of area $2$, so each segment is $\pi - 2$. Subtracting both: $2\pi - 2(\pi - 2) = 4$\dots so the shaded area is $4$.''}

\textbf{No tool calls were emitted.}

\textbf{\answer} \texttt{4} \(\times\)
\end{grpobox}

Both rollouts begin with the same algebraic derivation that subtracts the two overlapping segments from the top semicircle and arrives at $4$, the area of the wrong sub-region. The error is geometric, not arithmetic. AXPO emits a python call to render the configuration; the matplotlib output makes the area-swap structure visible -- the bottom quarter-disks at corners $E,F$ inside the rectangle (total area $2\pi$) match the top half-disk hanging from $AB$ (also $2\pi$), so the shaded region rearranges to the rectangle area $8 = 4 \times 2$, and the model revises accordingly. All four GRPO rollouts emit zero tool calls and commit to one of the wrong values ($4$, $4$, $4$, $2\pi$). The example illustrates a clean instance of the \gap: the algebraic prefix is sound but the area-swap intuition only becomes accessible after a tool call, and GRPO never learns to invoke it on questions of this type.

\subsection{Search: multi-hop decomposition into two sequential queries}
\label{app:qual-search}

\begin{questionbox}
\textbf{Benchmark}: HR-MMSearch~\citep{hrmmsearch}. \textbf{Ground truth}: \texttt{1984}.

\textbf{Prompt}: ``In which year was the mayor, who will be serving in August 2025, born in the city where the fashion designer of this fashion show was born?''
\begin{center}
\includegraphics[width=0.4\linewidth]{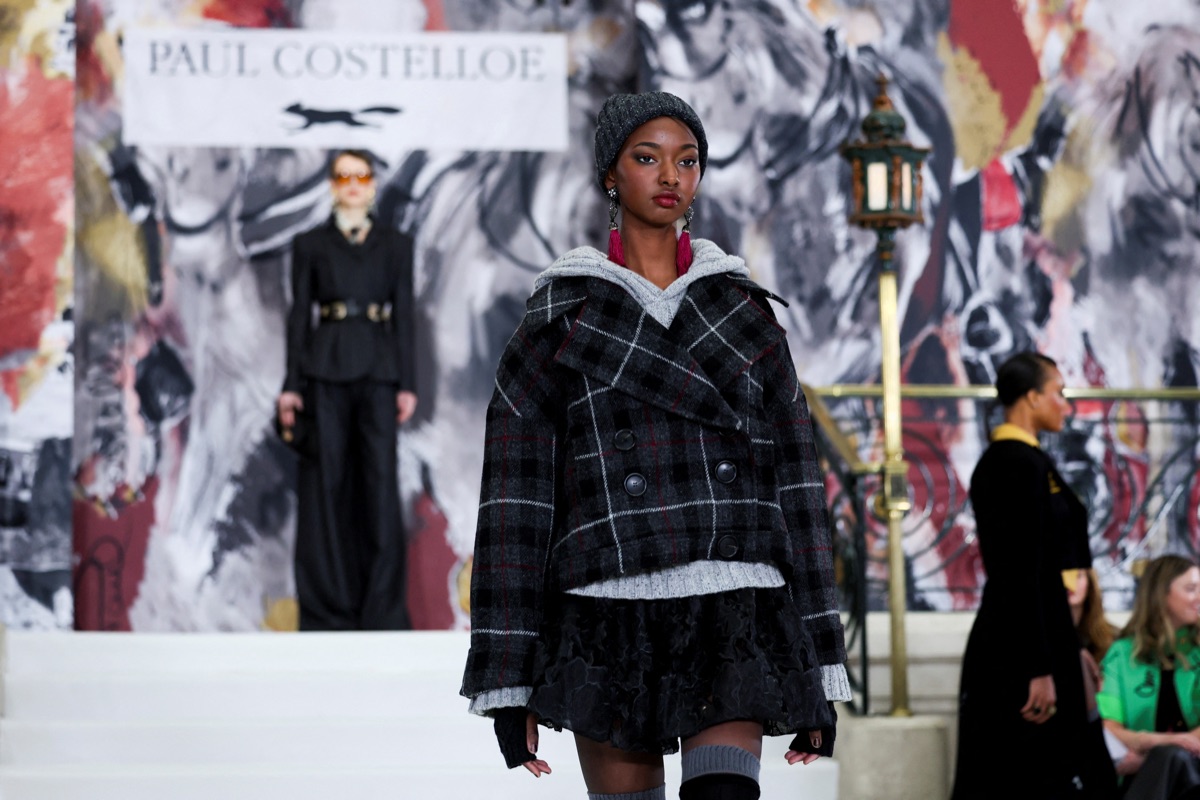}
\end{center}
\end{questionbox}

\begin{axpobox}
\textbf{Step 0, \think.} \emph{``The backdrop reads PAUL COSTELLOE; designer is Paul Costelloe. I need (i) his birthplace, (ii) the mayor of that city in August 2025, (iii) the mayor's birth year. Start with the birthplace.''}

\textbf{Step 0, \toolcall.}\\
{\small\ttfamily tavily-search(\{"query": "Paul Costelloe birthplace", "max\_results": 5\})}

\textbf{Tool output (truncated).} ``\emph{\dots Born in Dublin, Ireland \dots Paul Costelloe (23 June 1945 -- 21 November 2025)\dots}''

\textbf{Step 1, \think.} \emph{``Birthplace = Dublin. Now the Lord Mayor of Dublin in August 2025.''}

\textbf{Step 1, \toolcall.}\\
{\small\ttfamily tavily-search(\{"query": "current mayor of Dublin Ireland 2024", "max\_results": 5\})}

\textbf{Tool output (truncated).} ``\emph{Ray McAdam (born 1984) is an Irish Fine Gael politician who has served as Lord Mayor of Dublin since June 2025.}''

\textbf{Step 2, \think.} \emph{``Designer was born in Dublin; Lord Mayor of Dublin since June 2025 is Ray McAdam, born 1984.''}

\textbf{\answer} \texttt{1984} \(\checkmark\)
\end{axpobox}

\begin{grpobox}
\textbf{Step 0, \think.} \emph{``Designer is Paul Costelloe; born in Dublin in 1945. Mayor of Dublin in 2025 -- the search results don't directly show this, but if the riddle is that the mayor is the same age as the designer, the answer would be 1945.''}

\textbf{Step 0, \toolcall.}\\
{\small\ttfamily tavily-search(\{"query": "Paul Costelloe birthplace", \dots\})}

\textbf{Tool output (truncated).} ``\emph{\dots Born in Dublin, Ireland \dots Paul Costelloe (23 June 1945 -- 21 November 2025)\dots}''

\textbf{Step 1, \think.} \emph{``\dots search returned the designer's info but not the mayor's. I'll commit to 1945 since the designer was 80 in 2025 and the question may be a riddle.''}

\textbf{No second-hop search.}

\textbf{\answer} \texttt{1945} \(\times\)
\end{grpobox}

The two-hop structure (designer $\to$ city; city + date $\to$ mayor; mayor $\to$ birth year) admits a single-query collapse only by accident; the load-bearing decision is whether to issue a second \texttt{tavily-search} after the first hop returns. AXPO consistently issues the follow-up; all four GRPO rollouts on this question stop after the first hop and fall back to a riddle-style guess (1945, 1963, 1968, etc.).

\end{document}